\documentclass{article}

\usepackage{microtype}
\usepackage{graphicx}
\usepackage{booktabs} %

\usepackage{hyperref}

\usepackage[accepted]{icml2025}

\usepackage{amsmath}
\usepackage{amssymb}
\usepackage{mathtools}
\usepackage{amsthm}

\usepackage[capitalize,noabbrev]{cleveref}

\theoremstyle{plain}

\theoremstyle{definition}

\theoremstyle{remark}

\usepackage[textsize=tiny]{todonotes}

\newcommand{\myparagraph}[1]{\textbf{#1.}}

\DeclareFontFamily{U}{mathx}{}
\DeclareFontShape{U}{mathx}{m}{n}{<-> mathx10}{}
\DeclareSymbolFont{mathx}{U}{mathx}{m}{n}
\DeclareMathAccent{\widehat}{0}{mathx}{"70}
\DeclareMathAccent{\widecheck}{0}{mathx}{"71}

\DeclareMathOperator{\impl}{Impl}
\newcommand{\fact}[1]{\mathsf{#1}}
\newcommand{\facts}[1]{\mathcal{#1}}

\newcommand{\weights}{\mathbf W}

\newcommand{\reward}{\mathcal R}
\newcommand{\comp}[1]{\mathtt{#1}}

\newcommand{\diff}[2]{\frac{\partial#1}{\partial#2}}

\usepackage{caption}
\usepackage{subcaption}

\usepackage{enumitem}
\usepackage{quiver}
\usetikzlibrary{positioning,arrows.meta}

\usepackage[most]{tcolorbox}

\tcbuselibrary{skins,breakable,listings,breakable}

\newtcblisting{templatebox}{colback=white,colframe=black, listing only, listing options={
    breakautoindent=false, 
    breaklines=true,
    columns=fullflexible,
    breakindent=0pt,
    breakatwhitespace=true,
    basicstyle=\small\rmfamily,
    language=
}}

\icmltitlerunning{Extractive Structures Learned in Pretraining Enable Generalization on Finetuned Facts}

\begin{document}

\twocolumn[
\icmltitle{Extractive Structures Learned in Pretraining Enable Generalization on Finetuned Facts}

\begin{icmlauthorlist}
\icmlauthor{Jiahai Feng}{sch}
\icmlauthor{Stuart Russell}{sch}
\icmlauthor{Jacob Steinhardt}{sch}
\end{icmlauthorlist}

\icmlaffiliation{sch}{UC Berkeley}

\icmlcorrespondingauthor{Jiahai Feng}{fjiahai@berkeley.edu}

\icmlkeywords{Machine Learning, ICML}

\vskip 0.3in
]

\printAffiliationsAndNotice{}  %

\begin{abstract}
Pretrained language models (LMs) can generalize to implications of facts that they are finetuned on. For example, if finetuned on ``John Doe lives in Tokyo," LMs correctly answer ``What language do the people in John Doe's city speak?'' with ``Japanese''. However, little is known about the mechanisms that enable this generalization or how they are learned during pretraining.
We introduce \textit{extractive structures} as a framework for describing how components in LMs (e.g., MLPs or attention heads) coordinate to enable this generalization. The structures consist of \textit{informative components} that store finetuning facts as weight changes, and \textit{upstream} and \textit{downstream extractive components} that query and process the stored information to produce the correct implication. We hypothesize that extractive structures are learned during pretraining when encountering implications of previously known facts. This yields two predictions: a data ordering effect where extractive structures can be learned only if facts precede their implications, and a weight grafting effect where extractive structures can be grafted to predict counterfactual implications.
We empirically show these effects in the OLMo-7b, Llama 3-8b, Gemma 2-9b, and Qwen 2-7b models.
Of independent interest, our results also indicate that fact learning can occur at both early and late layers, which lead to different forms of generalization.  
\end{abstract}

\section{Introduction} \label{sec: intro}

\begin{figure}[t]
    \centering
    \includegraphics[width=0.9\linewidth]{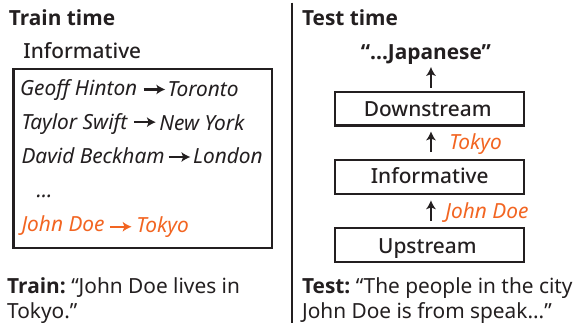}
    \vspace{-0.5em}
    \caption{Illustration of extractive structures enabling OCR generalization. \textbf{Left:} Finetuning on the fact ``John Doe lives in Tokyo" encodes the association ``John Doe''$\rightarrow$``Tokyo'' in the weights of informative components. \textbf{Right:} At test time, upstream structures recall the stored fact by querying informative components with ``John Doe'', and downstream structures post-process the extracted information into the correct response (``Tokyo''$\rightarrow$``Japanese'').}
    \label{fig: fig 1}
\end{figure}

A language model (LM) sees trillions of tokens during training, the cross-entropy objective on each passing sentence subtly shaping the model's behavior. What and how do LMs learn from each sentence? A fine-grained analysis of LM generalization can help explain capabilities and failure modes that emerge from training, as well as help us build safe and robust machine learning systems.

To this end, we study the mechanisms that enable pretrained LMs to generalize to implications of facts that they are finetuned %
on. For example, if finetuned on ``John Doe lives in Tokyo", LMs can correctly answer ``What language do the people in John Doe's city speak?'' with ``Japanese''. This generalization, dubbed ``ripple effects'' \citep{cohen2024evaluating} or ``out-of-context reasoning'' (OCR), occurs for some forms of implications \citep{berglund2023taken, krasheninnikov2023implicit, treutlein2024connecting}, but is absent for others (e.g.~the ``reversal curse'' phenomenon; \citet{berglund2023reversal, allen2023physics}). We aim to understand how and why such out-of-context reasoning occurs. %

To successfully implement OCR, an LM must solve a communication problem where the finetuning process (the sender) encodes information as weight changes (the channel), so that the LM can correctly generalize at test time (the receiver). Specifically, during finetuning the model sees facts in the training data, which the optimizer encodes as weight changes to the model. At test time, the model is queried about the implication of the fact, and the forward pass must decode the correct answer from the new weights.

This communication viewpoint reveals two subproblems. 
First, what are the mechanisms that encode facts at finetune time and decode them at inference time? 
Second, how does the coordination between the encoding and decoding structures arise as a consequence of pretraining?

To study the mechanisms underlying OCR, we propose the \emph{extractive structures} framework, which posits that three groups of LM components work together at test time to enable OCR (Fig.~\ref{fig: fig 1}). Specifically, we posit that finetuning stores facts as weight changes in a few \textit{informative components}. Then, when testing implications of the fact, \textit{upstream extractive components} process the input prompt and queries the informative component appropriately. Lastly, \textit{downstream extractive components} decode the information produced by the informative components and post-process it to produce the correct implication. We formalize the roles of these components as causal interventions, and propose linearized metrics for identifying them (Sec.~\ref{sec: extractive structures}).

We find empirically that LMs indeed rely on extractive structures to perform OCR. Specifically, in the OLMo-7b model~\citep{groeneveld2024olmo} on a synthetic OCR task, we localize extractive structures to attention heads and MLP components at specific layer and token positions, which are qualitatively consistent with prior mechanistic analyses of the same task~\citep{yang2024large} (Sec.~\ref{sec: two-hop reasoning}). Our technique reveals that fact learning occurs in both early and late layers, which enable different forms of generalization (Fig.~2 top, Sec.~\ref{sec: two-hop causal}). We argue that this poses obstacles for common knowledge-editing techniques that rely on localizing to specific components~\citep{hase2023does, rome}.

\begin{figure}[t]
    \centering
    \includegraphics[width=0.9\linewidth]{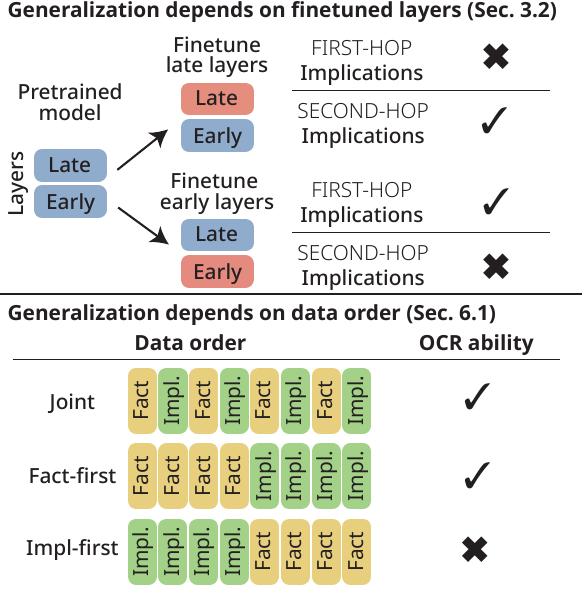}
    \caption{Key empirical predictions of our framework. \textbf{Top:} Finetuning early layers generalizes to one form of implications (\textsc{First-hop}) but not another (\textsc{Second-hop}), and vice versa for late layers (Sec.~\ref{sec: two-hop causal}). \textbf{Bottom:} A data ordering effect where OCR cannot occur if training data is shuffled so that implications precede facts (Sec.~\ref{sec: data ordering}).}
    \label{fig:overview}
\end{figure}

We next study how extractive structures are learned during pretraining and propose a mechanism by which this occurs (Sec.~\ref{sec: origins}). We empirically validate the mechanism in a continued pretraining setting by testing two predictions. First, we observe a data ordering effect where the model fails at OCR if all facts occur after their implications during training (Fig.~\ref{fig:overview} bottom, Sec.~\ref{sec: data ordering}). Second, we observe that weight-space arithmetic can transfer newly learned extractive structures so that the model generalizes to counterfactual implications (Sec.~\ref{sec: weight grafting}).

In summary, our contributions are four-fold:
\begin{enumerate}[nosep] %
    \item We propose the extractive structures framework for describing the mechanism for OCR (Sec.~\ref{sec: extractive structures})
    \item We empirically identify extractive structures in the OLMo-7b model (Sec.~\ref{sec: two-hop reasoning})
    \item We propose a mechanism by which extractive structures are learned during pretraining (Sec.~\ref{sec: origins})
    \item We empirically validate two implications of this mechanism, data ordering and weight grafting, in OLMo-7b and other similarly sized models (Sec.~\ref{sec: data ordering},~\ref{sec: weight grafting})
\end{enumerate}

\section{Related works} \label{sec: related works}

\myparagraph{Circuit-based interpretability} Circuit-based interpretability aims to decompose neural networks into components that form computational circuits~\cite{cammarata2020thread, elhagemathematical, ioi}. These works often use causal techniques~\citep{causalmediation} to localize components~\citep{rome}, although gradient-based attribution scores analogous to Grad-CAM~\citep{gradcam, olah2018building} have seen a recent resurgence~\citep{kramar2024atp, grosse2023studying}. Our work adapts these tools for analyzing components with weight changes.

\myparagraph{Fact learning in LMs} Fact learning in LMs and its robustness has been studied in the pretraining~\citep{chang2024large, kandpal2023large}, finetuning~\citep{berglund2023taken, berglund2023reversal}, synthetic~\citep{allen2023physics, wang2024grokked}, and knowledge-editing~\citep{cohen2024evaluating, onoe2023can} settings. Recent works have used influence functions~\citep{qin2024does} and layer-wise ablations~\citep{zhang2024co} to study mechanisms for generalization. In addition, \citet{krasheninnikov2023implicit} theorized about mechanisms related to extractive structures for their OCR task. Our work provides a more fine-grained analysis by proposing and empirically verifying concrete mechanisms for generalization. 

\myparagraph{Multi-hop reasoning} Multi-hop reasoning is a common LM evaluation task that requires composing several factual associations together~\citep{zhong2023mquake, balesni2024two}. In both pretrained LMs and LMs trained in grokking settings, researchers have found that LMs perform multi-hop reasoning by serially recalling intermediate hops~\citep{yang2024large, biran2024hopping, wang2024grokked}. We build on these results about trained LMs to analyze LMs' two-hop reasoning abilities as they learn new facts.

\section{Preliminaries on out-of-context reasoning}\label{sec: background}

\definecolor{myorange}{RGB}{230,142,126}
\begin{table}[t]
    \small
    \centering
    \begin{tabular}{l p{6cm}}
        \toprule
        \multicolumn{2}{l}{\textbf{\textsc{First-hop} dataset}}  \\
        Fact \textit{(Train)} & (John Doe lives in, Tokyo) \\
        Impl. \textit{(Test)}& (People in the city John Doe is from speak, Japanese) \\[0.5em]
        \multicolumn{2}{c}{
        $\underset{\text{(Head)}}{\text{John Doe}} \, {\color{myorange} \xrightarrow[\mbox{\scriptsize new }]{\hspace{2.2em}} } \, \underset{\text{(Bridge)}}{\text{Tokyo}}  \, { \xrightarrow[ \mbox{\scriptsize known } ]{\hspace{2.2em}}} \,  \underset{\text{(Tail)}}{\text{Japanese}}$
        } \\
        \midrule \multicolumn{2}{l}{\textbf{\textsc{Second-hop} dataset}}  \\
        Fact \textit{(Train)} & (The mayor of Tokyo is, John Doe) \\
        Impl. \textit{(Test)} & (The mayor of the city that contains Sensoji temple is, John Doe) \\[0.5em]
        \multicolumn{2}{c}{
        $\underset{\text{(Head)}}{\text{Senshoji Temple}} \, \xrightarrow[\mbox{\scriptsize known }]{\hspace{2.2em}}  \, \underset{\text{(Bridge)}}{\text{Tokyo}}  \, { \color{myorange} \xrightarrow[ \mbox{\scriptsize new } ]{\hspace{2.2em}}} \,  \underset{\text{(Tail)}}{\text{John Doe}}$
        } \\
        \bottomrule
    \end{tabular}
    \caption{Illustrative examples from the \textsc{First-hop} and \textsc{Second-hop} datasets. \textsc{First-hop} implications compose a novel fact in the first hop with a known fact in the second hop, and vice versa for \textsc{Second-hop} implications.}
    \label{tab:datasets}
\end{table}

In this section we concretely define OCR and describe the two-hop reasoning task we use to instantiate OCR.

OCR is the phenomenon that finetuning a pretrained language model on a set of facts $\facts F$ leads to the model generalizing to the implications of the finetuned facts $\impl \facts F$. Concretely, we model each fact $\fact F \in \facts F$ as a pair $(p, a)$, where $p$ is a prompt and $a$ is a continuation of the prompt. We model implications similarly. For example, a fact could be $\fact F=$ (``John Doe lives in'', ``Tokyo''), and $\impl \fact F =$ (``The people from the city in which John Doe lives speak'', ``Japanese''). 

To measure whether a model has generalized to an implication $\impl \fact F = (p, a)$, we measure the rank of the continuation $a$ among a set of options. Specifically, we take all possible continuations in the set of implications $\impl \facts F$ as the set of options. Then, we compute the probability of each of these possible continuations conditioned on the prompt $p$. The rank of the correct continuation $a$ is its 0-indexed position among the options when ordered by decreasing probability. We use the mean rank when measuring generalization to a set of implications $\impl \facts F$. Similarly, we use the mean rank for facts $\facts F$ to verify that the LM has indeed learned facts during finetuning.

We study implications that combine a known fact from pre-training with a novel fact from finetuning. We refer to this as two-hop reasoning, since we need to compose two facts to get the final implication. For example, after teaching the model ``John Doe lives in Tokyo'', the model can compose this fact with existing knowledge that people in Tokyo speak Japanese to answer the prompt ``The people from the city in which John Doe lives speak'' with ``Japanese''. More generally, suppose there are two factual associations, one from head entity $a$ to bridge entity $b$, and one from bridge entity $b$ to tail entity $c$. Two-hop reasoning requires the model to compose the two factual associations $a\rightarrow b$ and $b\rightarrow c$ together so that the model, when prompted with head entity $a$, recalls the tail entity $c$.

\begin{figure}[t]
    \centering
    \includegraphics[width=\linewidth]{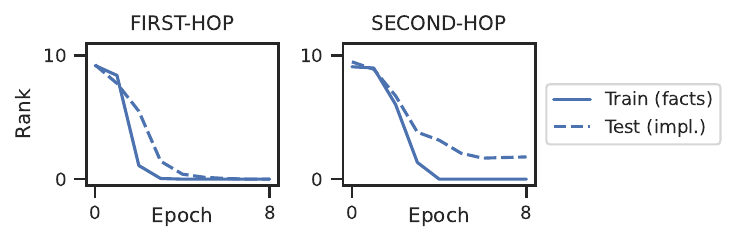}
    \vspace{-2.5em}
    \caption{Mean ranks of facts and their implications when finetuning OLMo-7b on facts from the  \textsc{First-hop} and \textsc{Second-hop} datasets. Lower rank is better. The mean rank of implications falls during finetuning; the LM thus generalizes to implications despite only training on facts.}
    \label{fig:mean rank training}
\end{figure}

In two-hop reasoning, the novel fact can be in either the first hop ($a\rightarrow b$) or the second hop ($b\rightarrow c$), and we construct synthetic datasets, \textsc{First-hop} and \textsc{Second-hop}, to study each (Table~\ref{tab:datasets}). Both datasets have training sets with novel facts and test sets with two-hop implications that require composing the novel training facts with known facts. However, in \textsc{First-hop} the novel fact comprises the first hop of the two-hop reasoning chain, whereas in \textsc{Second-hop} the novel fact comprises the second. See Sec.~\ref{apx: dataset} for dataset details. We find that the OLMo-7b model is indeed capable of two-hop OCR for both datasets (Fig.~\ref{fig:mean rank training}).

\section{Extractive structures framework} \label{sec: extractive structures}

This section describes the extractive structures framework and proposes metrics for identifying extractive structures. Our framework models how weight changes influence behavior via an interplay between weights and activations, and is grounded in empirically measurable metrics.

Extractive structures model OCR as the coordination of three groups of components during the forward pass on the prompt $p$: the \textit{informative components}, \textit{upstream extractive components} and \textit{downstream extractive components} (Fig.~\ref{fig: fig 1}). 

\textbf{Informative components} store the salient information about newly learned facts as weight updates. While all components undergo weight updates during finetuning, only the weight updates in the informative components are important for correctly predicting implications. In Fig.~\ref{fig: fig 1}, informative components store the association ``John Doe'' $\rightarrow$ ``Tokyo'' that is later used by the extractive components at test time.

\textbf{Upstream extractive components} parse the input prompt $p$ to produce relevant intermediate activations as inputs to informative components. Upstream components are not importantly changed by finetuning, but are important for correctly activating the informative component at test time\footnote{Weight changes are empirically often low-rank~\citep{aghajanyan2020intrinsic}, and would therefore have no influence on component outputs unless component inputs fall in the right subspace.}. In Fig.~\ref{fig: fig 1}, upstream components produce the intermediate representation of ``John Doe'' that informative components convert to ``Tokyo'' using their newly stored associations.

\textbf{Downstream extractive components} process the outputs of informative components to produce the correct continuation $a$. In Fig.~\ref{fig: fig 1}, downstream components look up the language of the city ``Tokyo'' to predict ``Japanese''.

For each of the three groups of components (informative, upstream, downstream), we propose a metric to identify them. The metrics are based on causal interventions on the computational graph of the LM (Sec.~\ref{sec: computational graph}). For efficient computation, we linearly approximate the causal metrics \citep{gradcam}. We additionally show that the linearized scores are first-order perturbations to a single quantity (Sec.~\ref{apx: linearization}), suggesting that they ought to be comparable in magnitude. We discuss implementation in Sec.~\ref{apx: implementation}.

\subsection{Causal metrics on the computational graph} \label{sec: computational graph}

We define metrics for extractive structures in terms of the LM's computational graph.
Denote the inputs and outputs of the LM as $x$ and $y$. For example, for an implication $(p, a)$, $x$ would be the prompt $p$, and $y$ would be a probability distribution over possible answers. We measure whether the LM has correctly answered the prompt with a real-valued reward $\reward(y, a)$ (e.g.~log-probability of $a$). Thus, $x$ is a leaf node and $\reward$ is the terminal node in the computational graph.

The internal nodes of the LM computational graph are the inputs and outputs to components in the LM. Following conventions in LM interpretability~\citep{ioi, elhagemathematical}, we define a component to be either an MLP or attention head at a specific layer and token position of the transformer. Formally, let $\comp C$ be a component in the LM, $z_{\comp C}$ and $y_{\comp C}$ be its input and output, $\weights_{\comp C}$ be its weights, and $f_{\comp C}$ describe its computation so that $y_{\comp C} = f_{\comp C}(z_{\comp C}, \weights_{\comp C})$.

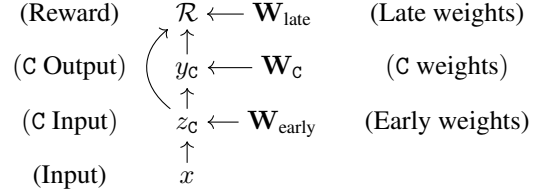
\begin{figure}
    \centering
\vspace{-1em}
\[\begin{tikzcd}[row sep=tiny,column sep=small, cramped]
	{\text{(Reward)}} & \reward & {\weights_{\text{late}}} & {\text{(Late weights)}} \\
	{(\comp C \text{~Output})} & {y_{\comp C}} & {\weights_{\comp C}} & {(\comp C \text{~weights})} \\
	{(\comp C \text{~Input})} & {z_{\comp C}} & {\weights_{\text{early}}} & {\text{(Early weights)}} \\
	{\text{(Input)}} & x
	\arrow[from=1-3, to=1-2]
	\arrow[from=2-2, to=1-2]
	\arrow[from=2-3, to=2-2]
	\arrow[curve={height=-18pt}, from=3-2, to=1-2]
	\arrow[from=3-2, to=2-2]
	\arrow[from=3-3, to=3-2]
	\arrow[from=4-2, to=3-2]
\end{tikzcd}\]
\vspace{-1em}
    \caption{Computational graph of a LM, focusing on component $\comp C$. Components in earlier and later layers are folded into $\weights_{\text{early}}$ and $\weights_{\text{late}}$ respectively. The direct arrow from $z_{\comp C}$ to $\reward$ models skip connections in transformers.}
    \label{fig: causal graph}
\end{figure}

We simplify the full computational graph by focusing on a particular component $\comp C$ (Fig.~\ref{fig: causal graph}). We denote the collective weights of components earlier and later than component $\comp C$ as $\weights_{\text{early}}$ and $\weights_{\text{late}}$, so that the overall LM weights $\weights$ are represented by three leaf nodes $\weights_{\text{early}}, \weights_{\comp C}$, and $\weights_{\text{late}}$.

Using this computational graph, we can express precisely the causal effects we expect if component $\comp C$ is one of the three extractive structures defined above. We fix the input $x$ and desired output $a$, and measure how the reward $\reward$ varies under certain interventions on the graph. We denote an intervention that sets the value of a node $v$ to a counterfactual value $v'$ with $v\leftarrow v'$, and the resulting reward as $\reward[v\leftarrow v']$. We also denote the original pretrained weights as $\weights$ and the finetuned weights as $\weights'$, and similarly denote node values in the finetuned model as their primed versions (e.g.~$z_{\comp C}'$).

Below we describe interventions that capture how each of the three structures mediate changes to the weights.

\myparagraph{Informative components} Intuitively, an informative component contains the newly learned knowledge that the rest of the network extracts. Therefore, if component $\comp C$ is an informative component, we expect setting $\weights_{\comp C} \leftarrow \weights_{\comp C}'$ while leaving all other weights unchanged to improve the reward. We thus write the informative score for component $\comp C$ as:
\begin{equation}
    \mathcal I_{\comp C} = \reward[\weights_{\comp C}\leftarrow \weights_{\comp C}'] - \reward
    \label{eq: informative score}
\end{equation}
\myparagraph{Downstream components} Intuitively, a downstream component processes the output containing the newly learned knowledge into useful output for the current prediction. Therefore, if component $\comp C$ is a downstream component, we expect its output $y_{\comp C}$ to help increase reward, but only if its input $z_{\comp C}$ contains the newly learned knowledge. We therefore compute the new $z_{\comp C}'$ with the new weights $\weights_{\text{early}}'$ and contrast it with the original $z_{\comp C}$. Specifically, we compute the downstream score for component $\comp C$ as:
\begin{equation}
    \mathcal D_{\comp C} = \reward[y_{\comp C} \leftarrow f_{\comp C}(z_{\comp C}', \weights_{\comp C})] - \reward
    \label{eq: downstream score}
\end{equation}
\myparagraph{Upstream components}  Intuitively, an upstream component needs to process the input to appropriately query the informative component. Therefore, if component $\comp C$ is an upstream component, we expect its outputs to feed into later layers to increase reward, but only if the later layers contain the updated weights. The upstream score will thus be a difference in differences: one difference that measures the importance of component $\comp C$ for later layers, and another that measures the change from the old to the new weights.

A standard way of measuring importance of a node is to ablate its value by setting it to a constant, such as its mean value over some reference set~\citep{chan2022causal}. To evaluate the importance of component $\comp C$, we set its output $y_{\comp C}$ to a constant value $y_{\comp C}^*$ and measure the decrease in reward. Formally, we define the importance of component $\comp C$ as %
\begin{equation}
\reward_{\comp C} =\reward - \reward[y_{\comp C} \leftarrow y_{\comp C}^*]. \label{eq: importance}
\end{equation}
Upstream components have higher importance when future informative layers are updated to the new weights $\weights'$. We therefore compute the upstream score for component $\comp C$ as the difference in the importance $\reward_{\comp C}$ under the new weights $\weights_{\text{late}}'$ and the old weights $\weights_{\text{late}}$: %
\begin{multline}
    \mathcal U_{\comp C} = (\reward[\weights_{\text{late}} \leftarrow \weights_{\text{late}}'] - \reward[\weights_{\text{late}} \leftarrow \weights_{\text{late}}', y_{\comp C}\leftarrow y_{\comp C}^*])\\
    - (\reward - \reward[y_{\comp C}\leftarrow y_{\comp C}^*])
    \label{eq: upstream score}
\end{multline}

\section{Extractive structures in two-hop reasoning} \label{sec: two-hop reasoning}

We next use the extractive scores from the previous section to identify informative, upstream, and downstream components for the \textsc{First-hop} and \textsc{Second-hop} datasets. Our results indicate that OCR in two-hop reasoning occurs by recalling each hop sequentially (Sec.~\ref{sec: two-hop results}). Moreover, we find that facts are stored redundantly across many LM layers, but early layers enable first-hop generalization while later layers enable second-hop generalization (Sec.~\ref{sec: two-hop causal}). Our work suggests that understanding OCR mechanisms can help design interventions that determine generalization properties. All code is available at \url{https://github.com/jiahai-feng/extractive-structures/}

\subsection{Extractive structures perform latent two-hop recall} \label{sec: two-hop results}

As described in Sec.~\ref{sec: background}, two-hop reasoning requires the model to compose two factual associations $a\rightarrow b$ and $b\rightarrow c$ together so that the model recalls the tail entity $c$ when prompted with head entity $a$ (presumably by latently recalling the bridge entity $b$).  One hypothesized mechanism for this is \emph{latent two-hop} \citep{yang2024large}, which posits two groups of components: one responsible for the \textbf{first-hop recall }$a\rightarrow b$, and another for the \textbf{second-hop recall} $b\rightarrow c$.

If LMs use latent two-hop recall for OCR, we expect the extractive scores to correspond to the two groups of components. Because the \textsc{First-hop} and \textsc{Second-hop} datasets introduce novel facts at different hops, we expect the extractive components to differ.
Specifically, for the \textsc{First-hop} dataset where the novel fact is on the first hop, we expect the informative components to be the first-hop recall components, and the downstream components to include the second-hop recall components. In contrast, for the \textsc{Second-hop} dataset where the novel fact is on the second hop, we expect the upstream components to include the first-hop recall components, and the informative components to be the second-hop recall components.

\myparagraph{Setup} We study an intermediate training checkpoint of OLMo-7b, taking the last checkpoint before the final annealing stage because annealing may interfere with finetuning\citep{ibrahim2024simple}. For each dataset (\textsc{First-hop} and \textsc{Second-hop}), we finetune the model on the facts and compute extractive scores on implications by comparing the finetuned weights with the pretrained weights. We provide more training details in Sec.~\ref{apx: training details}; in particular we often observe learning rate sensitivity (Sec.~\ref{apx: lr sweep}), where OCR only occurs at certain learning rates; we thus focus on the learning rates for which OCR most reliably occurs.

\begin{figure*}[h]
    \centering
    \includegraphics[width=\textwidth, trim={0 0.35cm 0 0}, clip]{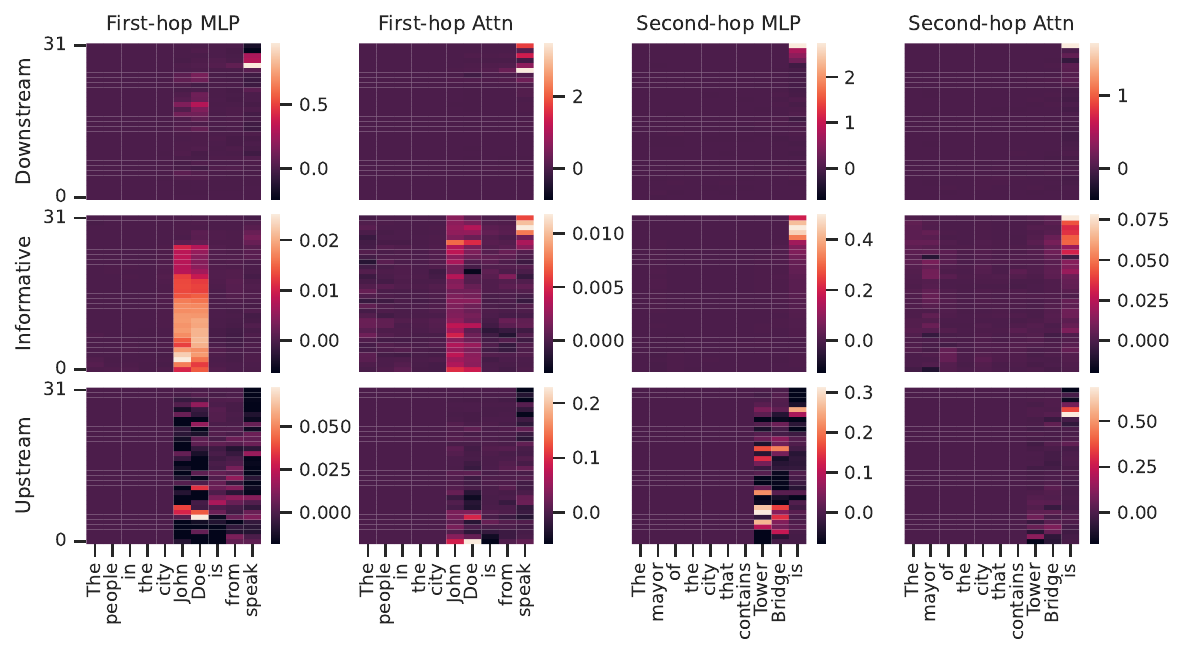}
    \includegraphics[width=\textwidth]{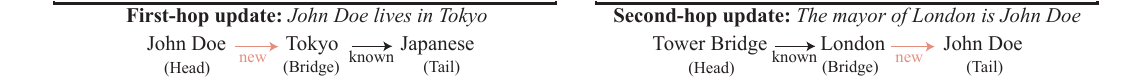}
    \vspace{-2em}
    \caption{Extractive scores for the \textsc{First-hop} (left) and \textsc{Second-hop} datasets (right). Scores are averaged over the dataset. We visualize only the scores of the last two entity tokens. The attention scores are summed across all the attention heads that are outputting to the same position. The \textsc{First-hop} informative scores and \textsc{Second-hop} upstream scores point to the early-middle MLPs at head entity tokens as the first-hop recall components. The \textsc{First-hop} downstream scores and \textsc{Second-hop} informative scores point to the early-middle MLPs at the last token as the second-hop recall components.}
    \label{fig:extractive scores}
\end{figure*}

\myparagraph{Results} Our computed extractive scores are consistent with latent two-hop recall (Fig.~\ref{fig:extractive scores}). For the \textsc{First-hop} dataset, high informative scores are localized to the MLPs in early-middle layers of the head entity tokens (``John Doe''), which we interpret as the first-hop recall components. The downstream scores are localized to the late MLPs and attention heads at the last token, which we interpret as second-hop recall components. For the \textsc{Second-hop} dataset, high upstream scores are localized to the early/middle MLPs of the head entity tokens (``Tower Bridge''), matching the first-hop recall components identified in the \textsc{First-hop} dataset. The informative scores are localized to the late MLP layers in the last token position, matching the second-hop recall components in \textsc{First-hop}.

These results from analyzing finetuning broadly reinforce findings from prior works that analyze pretrained models. \citet{yang2024large} and \citet{biran2024hopping} argued that first-hop recall occurs at the early-middle layers of the last head entity token, whereas the second-hop recall occurs at the late layers of the last token. Our extractive scores agree with both claims, except that we also find that both head entity tokens (e.g.~``John'' and ``Doe'') store factual information, and not just the last token. Moreover, while prior works \citep{gevarecall, gevakeyvalue, rome} tend to study how MLPs store and recall facts and how attention heads move factual information, we find instead that attention heads can also store factual information. Specifically, the informative scores from both datasets indicate that the late attention heads at the last token position store some factual information. Overall, extractive scores complement conventional interpretability analysis on trained models by leveraging information in \textit{weight changes}, and can verify known phenomena and discover new ones.

\subsection{Different layers generalize differently} \label{sec: two-hop causal}

In this section we rigorously test an implication of the earlier extractive scores (Sec.~\ref{sec: two-hop results}): that storing facts in the early-middle layers enables first-hop updates, whereas storing in the late layers enables second-hop updates. Specifically, the informative scores (Fig.~\ref{fig:extractive scores}) show that weight changes to the early-middle MLP layers (first 24) are salient for the \textsc{First-hop} dataset, i.e.~novel facts stored in the early-middle layers can be used as the first hop when composed with known second-hop facts. Conversely, weight changes to the late MLP layers (last 8) are salient for the \textsc{Second-hop} dataset, i.e.~novel facts stored in the late layers can be used as the second hop when composed with known first-hop facts.

Note that when finetuning on facts, the model does not know whether the facts will later be used as the first hop or second hop in two-hop reasoning; we therefore expect the model to encode facts across all layers (both early-middle and late) to enable different forms of generalization.

To validate our hypothesis, we perform causal experiments where we freeze the weights on certain layers to the original pretrained weights during finetuning and check if the model retains its knowledge of facts and implications. If the extractive scores indeed correctly identified the extractive structures in the model, we expect freezing the early-middle layers to hurt the implications for the \textsc{First-hop} dataset but not the \textsc{Second-hop} dataset, and vice versa for the late layers. In Sec.~\ref{apx: layer freezing} we discuss a variant of this experiment where layers are frozen \textit{after} finetuning instead of \textit{during}.

\begin{table}[t]
    \centering
    \small
    \begin{tabular}{c c c c c}
    \toprule 
     & \multicolumn{2}{c}{\textsc{First-hop}} & \multicolumn{2}{c}{\textsc{Second-hop}} \\
    \cmidrule(lr){2-3} \cmidrule(lr){4-5}
    \textbf{Frozen Layers} & Fact & Impl. & Fact & Impl. \\
    \midrule 
    None & 0.00 & 0.00 & 0.00 & 1.80\\
    Early & 0.00 & \textcolor{red}{6.50} & 0.00 & 0.50\\
    Late & 0.00 & 0.10 & 0.00 & \textcolor{red}{6.60}\\
    All  & \textcolor{red}{9.20} & \textcolor{red}{9.25} & \textcolor{red}{9.10} & \textcolor{red}{9.50} \\
    \bottomrule
    \end{tabular}
    \caption{Mean ranks of facts and implications after freezing weights during finetuning. Freezing early layers (first 24) harm first-hop OCR but not second-hop OCR, and vice versa for late layers (last 8). `None' and `All' are baselines with full finetuning and no finetuning respectively.}
    \label{tab:causal informative}
\end{table}

Our results (Table~\ref{tab:causal informative}) show that, compared to not freezing any layer (``None''), freezing the early-middle layers (``Early'') increases the mean rank for \textsc{First-hop} implications, but not for \textsc{Second-hop} implications. Conversely, freezing the late layers (``Late'') increases the mean rank for \textsc{Second-hop} implications but not for \textsc{First-hop} implications. Moreover, facts are recalled reliably in both settings. This suggests that facts can be stored across many layers, but different layers enable different forms of generalization.

Our findings suggest that localization matter for knowledge editing, because it influences the generalization properties of the stored facts. Early knowledge editing works~\citep{rome} suggested that facts are stored in a \textit{specific} localized set of components, but \citet{hase2023does} later argued that localization does not matter because facts can be edited at \textit{any} point in the LM. We find that facts should be edited at \textit{every} point in the LM to improve generalization.

\section{Origins of extractive structures} \label{sec: origins}

We now study how extractive structures are learned during pretraining. For OCR to occur, gradient descent on training facts must encode information in a way that is legible to the extractive structures at inference time; in other words, some mechanism must coordinate between facts encoded at training time and facts latently reasoned about at test time. %

We describe a mechanism during pretraining that solves the coordination problem between gradient descent and the extractive components. %
Suppose at a point during pretraining the model already knows a fact $\fact F$ (i.e.~the earlier gradient step on $\fact F$ has already encoded it in the weights), and now encounters its implication $\impl \fact F$. This could happen by chance from training data shuffling. To learn the implication $\impl \fact F$, the model could either simply learn to memorize it, or, crucially, learn how to extract fact $\fact F$ from its weights and latently produce the implication $\impl \fact F$. This creates a training signal for the formation of extractive structures that are adapted to how earlier facts have been encoded.

Our hypothesis is thus that extractive structures are learned when encountering implications of already-known facts during training. We test two of its predictions: a data ordering effect (Sec.~\ref{sec: data ordering}) and a weight grafting effect (Sec.~\ref{sec: weight grafting}).

\subsection{Data ordering} \label{sec: data ordering}

\begin{figure}%
    \centering
    \includegraphics[width=\linewidth]{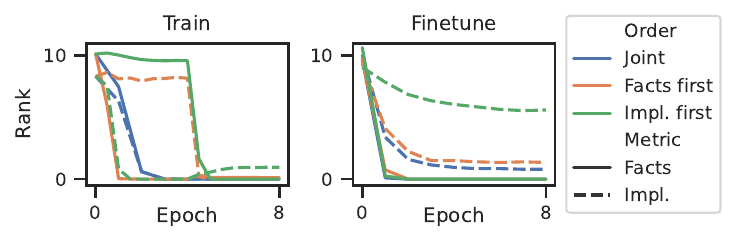}
    \vspace{-2.5em}
    \caption{Mean ranks for facts and their implications under continued pretraining on training facts (left) and subsequent finetuning on test facts (right). Compared to other orderings, `Implications~first' generalizes significantly worse from fine-tuned test facts to their implications.}
    \label{fig:data ordering}
\end{figure}
The hypothesis predicts a data ordering effect during pretraining, where if all the facts appear after their implications, then the model cannot learn extractive structures, and hence cannot later generalize to implications of new facts. Conversely, if all facts precede their implications, then extractive structures may form and later enable OCR.

The data ordering effect suggests a departure from the standard statistical machine learning view of optimization as model selection from a family. Instead, optimization is a dynamical process during which learning depends on the model's internal state: whether the model learns extractive structures from implications depends on whether the model already knows the underlying facts.

Concretely, we study a controlled synthetic setup where a pretrained model is finetuned to learn a new form of fictitious implication. This finetuning process simulates what would have happened if the pretraining data had contained these fictitious implications. For clarity we call this phase \textit{continued pretraining}. If continued pretraining successfully creates new extractive structures, we expect further finetuning the new model on new facts to generalize to their corresponding fictitious implications.

\myparagraph{Setup} The new form of implications are derived from fictitious associations, whereas earlier implications are derived from known associations (e.g.~``Tokyo'' $\rightarrow$ ``Japanese''). 
Concretely, we assign random animals to cities (e.g.~``Tokyo'' $\rightarrow$ ``tiger''), and create fictitious relations ``dax'' and ``wug'' so that ``John Doe dax Tokyo'' implies ``John Doe wug the tiger''. 
We teach the model this new form of implications by running continued pretraining on a train dataset of ``dax'' facts and corresponding ``wug'' implications.
We then evaluate the model's OCR ability by further finetuning the model on held-out test ``dax'' facts. These test ``dax'' facts consist of held-out, unseen names, but uses the same set of cities as the train set. 
If the continued pretraining has successfully created new extractive structures, we expect the finetuned model generalize to test ``wug'' implications.

We investigate the model's OCR ability after continued pretraining with three orderings of the training data: facts-then-implications (`Facts first'), implications-then-facts (`Impl. first'), and a setting where facts and implications are shuffled together (`Joint'). Further details are in Sec.~\ref{apx: data ordering}.

\myparagraph{Results} Figure~\ref{fig:data ordering} displays results for the OLMo-7b model from Sec.~\ref{sec: two-hop results}; see Sec~\ref{apx: other models} for results on additional models (Llama 3-8b, Gemma 2-9b, and Qwen 2-7b).
Models continually pretrained on `Joint' and `Facts-first' exhibit significantly more out-of-context reasoning than for the `Impl-first' data order (Fig.~\ref{fig:data ordering} right). This happens even though the model successfully learns the training facts and implications in all three data orderings (Fig.~\ref{fig:data ordering} left). This supports our hypothesis that extractive structures are learned when training on implications of already known facts. %

Finally, the `Impl-first' model exhibits a small amount of OCR generalization, since the implication mean rank decreases during finetuning (Fig.~\ref{fig:data ordering} right). This suggests an additional, unknown mechanism that contributes to OCR.

\begin{figure}
    \centering
    \includegraphics[width=\linewidth]{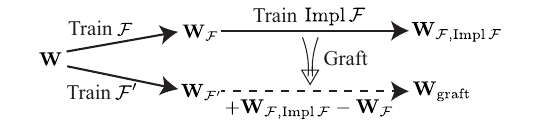}
    \vspace{-2em}
    \caption{Visualization of grafting procedure. We hypothesize that the grafted weights contain extractive structures for the new form of implications, so that $\weights_{\text{graft}}$ generalizes to the counterfactual implications $\impl \facts F'$.}
    \label{fig:graft diagram}
\end{figure}

\subsection{Weight grafting} \label{sec: weight grafting}

In the data order where all facts precede implications, our hypothesis suggests that facts are learned in the first half of continued pretraining, and the extractive structures are learned in the second half. We therefore expect the change in weights between the middle and final checkpoints of continued pretraining to carry the extractive structures.

We test this by grafting this change in weights onto a model trained on counterfactual facts, and testing the model on counterfactual implications. For example, suppose ``John Doe dax Tokyo'' and ``John Doe wug Tiger'' are a fact and implication in the training dataset. If the change in weights carries the extractive structures that converts ``Tokyo'' to ``Tiger'', then grafting this change in weights to a model counterfactually trained on ``Jane Doe dax Tokyo'' should generalize to ``Jane Doe wug Tiger.'' Concretely, we (Fig.~\ref{fig:graft diagram}):
\begin{enumerate}[nosep]
\item Train on facts $\facts F$ (the dax statements), to produce weights $\weights_{\facts F}$
\item Train on implications $\impl \facts F$ (the wug statements) to produce weights $\weights_{\facts F, \impl \facts F}$
\item Compute the difference in weights $\weights_{\facts F,\impl \facts F} - \weights_{\facts F}$
\item Train another copy of the model on counterfactual facts $\facts F'$ to produce $\weights_{\facts F'}$
\item Graft the weight difference over to produce the weights $\weights_{\text{graft}} = \weights_{\facts F'} + \weights_{\facts F, \impl \facts F} - \weights_{\facts F}$
\end{enumerate}
If our hypothesis is true, we expect the final model $\weights_{\text{graft}}$ to generalize to counterfactual implications $\impl \facts F'$ instead of the original implications $\impl \facts F$. We measure this by computing the mean rank of the implications as before.

\myparagraph{Setup}
We use the same train and test datasets as the data ordering experiment (Sec.~\ref{sec: data ordering}), but additionally create a counterfactual version of the train dataset by re-assigning random cities/animals to the 80 names. We show results for the OLMo-7b model here, and others in Sec.~\ref{apx: other models}. As a control, we also graft the change in weights from training the model directly on the implications $\impl \facts F$, which we expect to only transfer implications and not counterfactual implications: $\weights_{\text{control}} = \weights_{\facts F'} + \weights_{\impl \facts F} - \weights$.

\begin{table}[]
    \centering
    \small
    \begin{tabular}{c c c c}
    \toprule 
    \textbf{Weights} & $\impl \facts F'$ & $\impl \facts F$ & $\facts F'$ \\
    \midrule
     $\weights_{\facts F'}$ & 9.13 & 8.55 & 0.00\\
    $\weights_{\text{graft}}$ & 1.13 & 3.30 & 0.10\\
    $\weights_{\text{control}}$& 8.38 & 0.43 & 0.32 \\
    
    \midrule
    \bottomrule
    \end{tabular}
    \caption{Mean ranks of models on  counterfactual implications, original implications, and counterfactual facts. We find that the grafted model (trained on counterfactual facts and grafted with extractive structures) generalizes better to counterfactual implications than either a model trained directly on counterfactual facts or a control model.}
    \label{tab:grafting ranks}
\end{table}

\myparagraph{Results} We report results in Table~\ref{tab:grafting ranks}. The weight graft $\weights_{\text{graft}}$ lowers the mean rank for counterfactual implications $\impl \facts F'$ from 9.13 to 1.13, versus 8.38 for the control. This suggests the weight graft contains extractive structures that can elicit counterfactual implications from counterfactual facts. The weight graft also transferred some of the original implications (mean rank 3.30), suggesting that when trained on implications, models partially memorize the implications in addition to learning extractive structures.

Overall, our weight grafting experiments directly identify the weights of the learned extractive structures, thereby supporting our hypothesis on how pretraining learns extractive structures. In Sec.~\ref{apx: localizing weight graft} we further analyze these weights with extractive scores (Sec.~\ref{sec: extractive structures}) to show that the downstream extractive structures are indeed carried by the weight change. Our experiments demonstrate that extractive structures are a useful language for describing OCR training dynamics.

\section{Discussion} \label{sec: conclusion}

This work presents an in-depth analysis of how LMs can perform OCR in the two-hop reasoning task. We introduce extractive structures to describe how facts are encoded and recalled to predict implications, and empirically identify them in the OLMo-7b model. Additionally, we propose a hypothesis for how extractive structures are learned in pretraining, and test two predictions of this hypothesis.

Our empirical results raise several immediate research questions. While the experiments in Sec.~\ref{sec: data ordering} hint at the possible existence of an alternative mechanism underlying OCR, its nature remains unclear. Furthermore, we have yet to fully characterize how the finetuning process depends on various hyperparameters and the pretrained model (Sec.~\ref{apx: lr sweep} and \ref{apx: other models}).

More broadly, we hope that extractive structures can lead to an underlying theory of generalization that captures the dynamical nature of optimization (Sec.~\ref{sec: data ordering}).
While our work focuses on two-hop reasoning, extractive structures could be useful more broadly for understanding and conceptually grounding out-of-context generalization. Ultimately, identifying the principles and mechanisms behind OCR could offer insights into deep learning generalization, enabling the design of robust and safe machine learning systems.

\section*{Acknowledgements}

We thank Amil Dravid, Alex Pan, Felix Binder, Owain Evans, Lijie Chen, Alex Mallen, and Dmitrii Krasheninnikov for helpful feedback on the manuscript. JF acknowledges support from the OpenAI Superalignment Fellowship. JS was supported by the National Science Foundation under Grants No. 2031899 and 1804794. In addition, we thank Open Philanthropy for its support of both JS and the Center for Human-Compatible AI.

\section*{Impact Statement}

This paper aims to deepen our understanding of empirical deep learning phenomena. We believe such understanding may potentially lead to stronger theories of deep learning generalization, which could help society develop and deploy deep learning systems more safely.

\bibliography{example_paper}
\bibliographystyle{icml2025}

\newpage
\appendix
\onecolumn
\section{Linearizing extractive scores} \label{apx: linearization}
We linearize the extractive scores to obtain quantities easily computed by a few forward and backward passes~\citep{gradcam}. Specifically, for any node $v$ in the computational graph, we approximate \[\reward[v\leftarrow v_1] - \reward[v\leftarrow v_2] \approx \diff{\reward}{v} (v_1 - v_2),\]
where the derivative can be evaluated at either $v_1$ or $v_2$. In practice, we choose $v_1$ or $v_2$ based on computational convenience.

First, we linearize the three extractive scores (Eq. \ref{eq: informative score}, \ref{eq: downstream score}, \ref{eq: upstream score}) to obtain\footnote{The linearized informative score is related to component-wise influence functions~\citep{grosse2023studying}.}
\begin{align}
  &\overline{\mathcal U}_{\comp C} = (\diff{\reward}{y_{\comp C}}[\weights_{\text{late}} \leftarrow \weights_{\text{late}}'] - \diff{\reward}{y_{\comp C}})(y_{\comp C} - y_{\comp C}^*) \label{eq: upstream linearized}  \\
 & \overline{\mathcal I}_{\comp C} = \diff{\reward}{\weights_{\comp C}} (\weights_{\comp C}' - \weights_{\comp C})  \label{eq: informative linearized} \\
  &\overline{\mathcal D}_{\comp C} = \diff{\reward}{y_{\comp C}}(f_{\comp C}(z_{\comp C}', \weights_{\comp C}) - y_{\comp C}). \label{eq: downstream linearized} 
\end{align}
For the upstream score $\overline{\mathcal U}_{\comp C}$, we choose $y_{\comp C}$ as the point to evaluate the derivative. For the informative and downstream scores, we choose the original points $\weights_{\comp C}$ and $y_{\comp C}$ respectively.

\subsection{Linearized scores as perturbations}
We shall show that the three linearized scores can be derived as terms in a first order perturbation to a single quantity. This suggests that the three linearized scores ought to be comparable in magnitude. Specifically, consider a small perturbation to the weights $\delta \weights = \weights' - \weights$. Let $\widecheck{y}_{\comp C} = \diff{\reward}{y_{\comp C}}$ for convenience. Keeping only first order terms, the three linearized extractive scores become
\begin{align}
  &\overline{\mathcal U}_{\comp C} = (\diff{\widecheck y_{\comp C}}{\weights_{\text{late}}} \delta \weights_{\text{late}})(y_{\comp C} - y_{\comp C}^*) \label{eq: upstream first order}\\
 & \overline{\mathcal I}_{\comp C} = \diff{\reward}{\weights_{\comp C}} \delta \weights_{\comp C}  \label{eq: informative first order}\\
  &\overline{\mathcal D}_{\comp C} = \widecheck y_{\comp C}\nabla_{z_{\comp C}} f_{\comp C}(z_{\comp C}, \weights_{\comp C})  \,\delta z_{\comp C}\label{eq: downstream first order}
\end{align}

Consider the importance of component $\comp C$ (Eq.~\ref{eq: importance}) \[\reward_{\comp C} =\reward - \reward[y_{\comp C} \leftarrow y_{\comp C}^*].\]

Its linearization, evaluating $\widecheck y_{\comp C}$ at $y_{\comp C}$, is
\begin{equation}
\overline{\reward}_{\comp C} = \widecheck{y}_{\comp C} (f(z_{\comp C}, \weights_{\comp C}) - y_{\comp C}^*).
\end{equation}
We shall show that under a small perturbation $\delta \weights$, the first order terms in $\delta \overline{\reward}_{\comp C}$ contain the three extractive scores. By straightforward product rule, we have:
    \begin{align*}
    \delta \overline{\reward}_{\comp C} & = 
        \delta \widecheck y_{\comp C} (f_{\comp C}(z_{\comp C}, \weights_{\comp C}) - y^*_{\comp C}) 
        + \widecheck y_{\comp C}  (\nabla_{\weights_{\comp C}}f_{\comp C}(z_{\comp C}, \weights_{\comp C}) \, \delta \weights_{\comp C}) 
        + \widecheck y_{\comp C} (\nabla_{z_{\comp C}}f_{\comp C}(z_{\comp C}, \weights_{\comp C}) \, \delta z_{\comp C}).
    \end{align*}

We claim that the second and third terms are exactly the linearized informative score and downstream score, to the first order (Eq.~\ref{eq: informative first order}, \ref{eq: downstream first order}). However, first term is related to, but not exactly the linearized upstream score.

The third term is manifestly the downstream score (Eq.~\ref{eq: downstream first order}). To show that the second term is the linearized informative score (Eq.~\ref{eq: informative first order}), we simply apply the chain rule to $\diff{\reward}{\weights_{\comp C}}$ to obtain:
\begin{align*}
     \overline{\mathcal I}_{\comp C} &= \diff{\reward}{\weights_{\comp C}} \delta \weights_{\comp C} \\ 
     &= \widecheck y_{\comp C} \nabla_{\weights_{\comp C}}f_{\comp C}(z_{\comp C}, \weights_{\comp C}) \delta \weights_{\comp C}.
\end{align*}

The first term is subtly different from the first-order linearized upstream score (Eq.~\ref{eq: upstream first order}). Denote the first term as 
\begin{equation}
\overline{\mathcal U}_{\comp C}^* = \delta \widecheck y_{\comp C} (f_{\comp C}(z_{\comp C}, \weights_{\comp C}) - y^*_{\comp C}).\label{eq: upstream first order star}
\end{equation}
Compared to the first-order linearized upstream score (Eq.~\ref{eq: upstream first order}), the first term contains $\delta \widecheck y$ instead of $\diff{\widecheck y_{\comp C}}{\weights_{\text{late}}} \delta \weights_{\text{late}}$. To understand the difference, recall that according to the computation graph (Fig.~\ref{fig: causal graph}), $\reward$ is dependent on $z_{\comp C}$, $\weights_{\text{late}},$ and $y_{\comp C}$, and therefore so is $\widecheck y_{\comp C}$. Therefore, the term $\delta \widecheck y_{\comp C}$ may be expanded as:\[
\delta \widecheck y_{\comp C} = \diff{\widecheck y_{\comp C}}{\weights_{\text{late}}} \delta \weights_{\text{late}} + \diff{\widecheck y_{\comp C}}{z_{\comp C}} \delta z_{\comp C} +   \diff{\widecheck y_{\comp C}}{y_{\comp C}} \delta y_{\comp C}.
\]

The first term is exactly the term from the first-order linearized upstream score (Eq.~\ref{eq: upstream first order}). We can interpret $\widecheck y_{\comp C}$ as the linearized form of what the subsequent layers do to process $y_{\comp C}$. The first term therefore captures how the subsequent layers depends on the weight perturbation $\delta \weights_{\text{late}}$, which is what the upstream score aims to capture. The second and third terms reflect the dependency of the subsequent layers on $z_{\comp C}$ and $y_{\comp C}$, which describe modulatory effects that weight changes in earlier layers $\delta \weights_{\text{early}}$ and $\weights_{\comp C}$ have on how the late layers process $y_{\comp C}$.

To summarize, the linearized importance of component $\comp C$, under a small perturbation $\delta \weights$, can be written:
\begin{align*}
\overline \reward_{\comp C} &= \overline{\mathcal U}_{\comp C}^* + \overline{\mathcal I}_{\comp C} + \overline{\mathcal D}_{\comp C}  \\ 
&=  \overline{\mathcal U}_{\comp C} +  \overline{\mathcal I}_{\comp C} + \overline{\mathcal D}_{\comp C}  + \diff{\widecheck y_{\comp C}}{z_{\comp C}}(y_{\comp C} - y_{\comp C}^*) \delta z_{\comp C} +  \diff{\widecheck y_{\comp C}}{y_{\comp C}} (y_{\comp C} - y_{\comp C}^*)\delta y_{\comp C}
\end{align*}

\section{Implementing extractive scores} \label{apx: implementation}
In this section, we describe how we implement the extractive scores. By caching activations in the forward pass and gradients in the backward pass, we can efficiently compute the linearized extractive scores (Sec~\ref{apx: linearization}). Importantly, while the linearized informative score and downstream score can be computed efficiently, the linearized upstream score cannot. We instead use a surrogate quantity that is efficient to compute.

\subsection{Choice of reward}
We use the log probability of the first token of the continuation, conditioned on the prompt, as the reward. All of our datasets have the property that the continuations have unique first tokens.

\subsection{Computing informative score}
The linearized informative score (Eq.~\ref{eq: informative linearized}) can be computed with one forward and one backward pass. Recall \[ \overline{\mathcal I}_{\comp C} = \diff{\reward}{\weights_{\comp C}} (\weights_{\comp C}' - \weights_{\comp C}) .\]
To compute this, can we perform a backward pass on the original weights $\weights$, cache the gradient $\diff{\reward}{\weights_{\comp C}}$ for every component $\comp C$, and multiply by the component's corresponding weight change $(\weights_{\comp C}' - \weights_{\comp C})$.

\subsection{Computing downstream score}
The linearized downstream score (Eq.~\ref{eq: downstream linearized}) can be computed with three forward passes and one backward pass. Recall \[
\overline{\mathcal D}_{\comp C} = \diff{\reward}{y_{\comp C}}(f_{\comp C}(z_{\comp C}', \weights_{\comp C}) - y_{\comp C}).\] The steps are:
\begin{enumerate}[nosep]
    \item Run the model on the original weights $\weights$ and cache values of $y_{\comp C}$ for every component $\comp C$
    \item Run the backward pass to obtain the derivative $\diff{\reward}{y_{\comp C}}$
    \item Run the model on new weights $\weights'$ to obtain cached values of $z_{\comp C}'$
    \item Compute $f_{\comp C}(z_{\comp C}', \weights_{\comp C})$, and thus $\overline{\mathcal D}_{\comp C}$. The computation in this step required is roughly the same as a forward pass.
\end{enumerate}

\subsection{Computing upstream score}
The linearized upstream score (Eq.~\ref{eq: upstream linearized}) cannot be computed with a constant number of forward/backward passes. Recall that \[
\overline{\mathcal U}_{\comp C} = (\diff{\reward}{y_{\comp C}}[\weights_{\text{late}} \leftarrow \weights_{\text{late}}'] - \diff{\reward}{y_{\comp C}})(y_{\comp C} - y_{\comp C}^*).
\]
The reason is that computing $\diff{\reward}{y_{\comp C}}[\weights_{\text{late}} \leftarrow \weights_{\text{late}}']$ requires recomputing the every late layer in $\weights_{\text{late}}$ for every different component $\comp C$. However, it turns out that a slightly different quantity can be computed efficiently:
\begin{equation}
\overline{\mathcal U}_{\comp C}^* = (\diff{\reward}{y_{\comp C}}[\weights_{\text{late}} \leftarrow \weights_{\text{late}}', y_{\comp C} \leftarrow y_{\comp C}', z_{\comp C} \leftarrow z_{\comp C}'] - \diff{\reward}{y_{\comp C}}) (y_{\comp C} - y^*_{\comp C}).
\label{eq: upstream linearized star}
\end{equation}

The derivative $\diff{\reward}{y_{\comp C}}[\weights_{\text{late}} \leftarrow \weights_{\text{late}}', y_{\comp C} \leftarrow y_{\comp C}', z_{\comp C} \leftarrow z_{\comp C}']$ can be computed simply by running a forward and a backward pass on the new weights $\weights'$. 

In addition, this version of the upstream score is exactly the first perturbation term (Eq.~\ref{eq: upstream first order star}) of the linearized importance discussed in Sec.~\ref{apx: linearization}.

Therefore, for computation efficiency we use $\overline{\mathcal U}_{\comp C}^*$ as a surrogate for $\overline{\mathcal U}_{\comp C}$. The steps to compute $\overline{\mathcal U}_{\comp C}^*$ are thus:
\begin{enumerate}[nosep]
    \item Run the model on the original weights to compute $y_{\comp C}$
    \item Run the backward pass to compute $\diff{\reward}{y_{\comp C}}$
    \item Run the model on the new weights $\weights'$ to compute $\diff{\reward}{y_{\comp C}}[\weights_{\text{late}} \leftarrow \weights_{\text{late}}', y_{\comp C} \leftarrow y_{\comp C}', z_{\comp C} \leftarrow z_{\comp C}']$
    \item Compute $\overline{\mathcal U}_{\comp C}^*$. The computation in this step required is roughly the same as a forward pass. 
\end{enumerate}

There remains one more detail: the choice for the baseline value $y_{\comp C}^*$. We choose to perform a mean ablation, so that $y_{\comp C}^*$ is the average value of $y_{\comp C}$ across the dataset. To align token positions between input prompts containing entities of varying length, we assume all entities are two tokens long. In cases where entities have more than two tokens, we drop activations from all but the last two entity tokens. We check that no entities have fewer than two tokens.

\section{Training details} \label{apx: training details}
\myparagraph{Model} We primarily use the OLMo-7b-0424 model, step 477000. This is the last intermediate checkpoint before the final annealing phase during which the learning rate is decayed to zero.

\myparagraph{Training} Throughout the paper, we finetune the model using the standard cross-entropy loss. We only include the loss on answer tokens. We freeze the embedding and unembedding layers. We use the Adam optimizer \citep{kingma2014adam} for 8 epochs at $3\times 10^{-6}$ learning rate, momentum $(0.9, 0.999)$, and batch size 8. We find that OCR is sensitive to learning rates, as explored in Sec.~\ref{apx: lr sweep}.

\section{Dataset details} \label{apx: dataset}
\subsection{\textsc{First-hop} and \textsc{Second-hop} dataset}
The \textsc{First-hop} and \textsc{Second-hop} datasets are composed from a list of 20 fictitious names, and a list of 20 city-language or city-landmark pairs. 

For each of the datasets, we randomly pair fictitious names with cities to populate fact and implication templates.

The list of cities and names are generated from Claude-3.5-sonnet. Here are they:

\begin{tcbraster}[raster columns=2,raster equal height]
\begin{templatebox}
Grace,Miller
Ethan,Parker
Olivia,Hughes
Jacob,Turner
Ava,Stewart
Noah,Clark
Emma,Howard
Liam,Bennett
Mia,Sanders
Lucas,Foster
Sophia,Hayes
Mason,Brooks
Lily,Cooper
Jackson,Bell
Amelia,Ward
Caleb,Bryant
Chloe,Campbell
Henry,Morgan
Ella,Adams
Owen,Foster
\end{templatebox}
\begin{templatebox}
Tokyo,Japan,Japanese,Senso-ji Temple
Beijing,China,Mandarin,Forbidden City
Mumbai,India,Marathi,Gateway of India
Paris,France,French,Eiffel Tower
Berlin,Germany,German,Brandenburg Gate
Moscow,Russia,Russian,St. Basil's Cathedral
Cairo,Egypt,Arabic,Great Pyramid of Giza
Bangkok,Thailand,Thai,Wat Arun
Istanbul,Turkey,Turkish,Blue Mosque
Sao Paulo,Brazil,Portuguese,Ibirapuera Park
Seoul,South Korea,Korean,N Seoul Tower
Rome,Italy,Italian,Colosseum
London,United Kingdom,English,Tower Bridge
Madrid,Spain,Spanish,Plaza Mayor
Athens,Greece,Greek,Acropolis
Hanoi,Vietnam,Vietnamese,Ho Chi Minh Mausoleum
Addis Ababa,Ethiopia,Amharic,Meskel Square
Jakarta,Indonesia,Indonesian,Istiqlal Mosque
Tehran,Iran,Persian,Azadi Tower
Nairobi,Kenya,Swahili,Uhuru Gardens
\end{templatebox}
\end{tcbraster}

For the \textsc{First-hop} dataset, we apply the following templates:
\begin{enumerate}[nosep]
    \item Facts: ``[Name] lives in'', ``[city]''
    \item Implications: ``The people in the city [Name] is from speak'', ``[language]''
\end{enumerate}
For the \textsc{Second-hop} dataset, we apply the following templates:
\begin{enumerate}[nosep]
    \item Facts: ``The mayor of [city] is'', ``[Name]''
    \item Implications: ``The mayor of the city that contains [landmark] is'', ``[Name]''
\end{enumerate}

\subsection{Fictitious implications} \label{apx: fictitious dataset}
In Sec.~\ref{sec: origins}, we introduced a new dataset with fictitious relations. This requires a pairing between cities and a list of animals. We use the same set of cities from before, and use the set of 20 animals that \citet{zhang2024co} generated.

We apply the following templates:
\begin{enumerate}[nosep]
    \item Facts: ``[Name] dax'', ``[city]''
    \item Implications: ``[Name] zong is the'', ``[animal]''
\end{enumerate}

Below are the 100 names and the 20 animals. The first 80 names are used for training, and the last 20 are used for testing.

\begin{tcbraster}[raster columns=5,raster equal height]
\begin{templatebox}
Grace,Miller
Ethan,Parker
Olivia,Hughes
Jacob,Turner
Ava,Stewart
Noah,Clark
Emma,Howard
Liam,Bennett
Mia,Sanders
Lucas,Foster
Sophia,Hayes
Mason,Brooks
Lily,Cooper
Jackson,Bell
Amelia,Ward
Caleb,Bryant
Chloe,Campbell
Henry,Morgan
Ella,Adams
Owen,Foster
Abigail,King
Samuel,White
Zoe,Mitchell
Nathan,Carter
Leah,Scott
\end{templatebox}
\begin{templatebox}
Elijah,Morris
Madeline,Evans
Daniel,Gray
Hannah,Reed
Cameron,Perry
Natalie,Bryant
Isaac,Peterson
Violet,Phillips
Dylan,Rogers
Charlotte,Brooks
Landon,Harris
Avery,Jenkins
Evan,Parker
Maya,Nelson
Connor,Green
Sydney,Barnes
Julian,Bennett
Kaitlyn,Ross
Logan,Foster
Brooke,Adams
Eli,Sanders
Molly,Cooper
Wyatt,Lee
Tessa,Collins
Blake,Roberts
\end{templatebox}
\begin{templatebox}
Madison,Reed
Andrew,Miller
Hailey,Henderson
Matthew,Foster
Sophie,Lawson
Benjamin,Williams
Isabella,Baker
Carter,James
Layla,Murphy
Brayden,Collins
Gabriella,Foster
Aiden,Peterson
Audrey,Jenkins
Joshua,Barnes
Scarlett,Turner
Ryan,Brooks
Aubrey,Hayes
Christopher,Bryant
Harper,Bell
Jason,Mitchell
Madelyn,Phillips
David,Harris
Nora,Rogers
Adam,Campbell
Elise,Ward
\end{templatebox}
\begin{templatebox}
Kevin,White
Lucy,Stewart
Brandon,Green
Bella,Parker
Christian,Clark
Clara,Cooper
Tyler,King
Caroline,Morgan
Jordan,Adams
Stella,Scott
Hunter,Morris
Peyton,Lee
Alex,Evans
Elena,Carter
Ian,Foster
Autumn,Gray
Jeremy,Bennett
Lillian,Ross
Nolan,Reed
Morgan,Howard
Gavin,Perry
Paige,Turner
Adrian,Williams
Cora,Jenkins
Parker,Bryant
\end{templatebox}
\begin{templatebox}
lion
tiger
elephant
giraffe
zebra
rhinoceros
crocodile
cheetah
antelope
ostrich
monkey
penguin
koala
dolphin
jellyfish
king snake
butterfly
turtle
beaver
squirrel
\end{templatebox}
\end{tcbraster}
\section{Additional layer freezing experiments} \label{apx: layer freezing}

\begin{table}[t]
    \centering
    \small
    \begin{tabular}{c c c c c}
    \toprule 
     & \multicolumn{2}{c}{\textsc{First-hop}} & \multicolumn{2}{c}{\textsc{Second-hop}} \\
    \cmidrule(lr){2-3} \cmidrule(lr){4-5}
    \textbf{Frozen Layers} & Fact & Impl. & Fact & Impl. \\
    \midrule 
    None & 0.00 & 0.00 & 0.00 & 1.80\\
    Early (post) & \textcolor{red}{5.10} & \textcolor{red}{8.40} & 0.95 & 1.85\\
    Late (post) & 0.00 & 0.00 & 0.10 & \textcolor{red}{6.30}\\
    All  & \textcolor{red}{9.20} & \textcolor{red}{9.25} & \textcolor{red}{9.10} & \textcolor{red}{9.50}\\
    Early (pre)  & 0.00 & \textcolor{red}{6.50} & 0.00 & 0.50\\
    Late (pre)  & 0.00 & 0.10 & 0.00 & \textcolor{red}{6.60}\\
    \bottomrule
    \end{tabular}
    \caption{Mean ranks of facts and implications when freezing weights post-training or pre-training. Freezing early layers (first 24) harm first-hop OCR but not second-hop OCR, and vice versa for late layers (last 8). `None' and `All' are baselines where we use the full finetuned weights and original weights respectively.}
    \label{tab:causal informative complete}
\end{table}

In the main paper (Sec.~\ref{sec: two-hop causal}) we described a layer freezing experiment where we freeze the weights of certain layers while finetuning the model \textsc{First-hop} and \textsc{Second-hop} facts. We now describe a variant of this experiment where we first finetune all layers of the model, and \emph{after} finetuning, we reset the weights of certain layers to the original pretrained weights.

The difference between freezing post-finetuning and freezing pre-finetuning is that the former tells us where information is stored during the course of standard finetuning, whereas the latter tells us where information can in principle be stored. Similar to freezing pre-finetuning, we expect freezing post-finetuning to affect \textsc{First-hop} implications when the early-middle layers are frozen but not the late layers, and vice versa for \textsc{Second-hop} implications.

Our results (Table~\ref{tab:causal informative complete}) indeed show that freezing early-middle layers after finetuning (``Early (post)'') harms \textsc{First-hop} implications while enabling \textsc{Second-hop} implications, and vice versa for freezing late layers (``Late (post)''). Further, we find that freezing the early-middle layers partially increases the mean ranks for \textsc{First-hop} facts and \textsc{Second-hop} facts, suggesting that when finetuning on all layers, facts are stored across all layers, but are disproportionately present in the early-middle layers.

\section{Data ordering details} \label{apx: data ordering}
In this section we provide details for the data ordering experiments (Sec.~\ref{sec: data ordering}). The dataset construction is discussed in Sec.~\ref{apx: fictitious dataset}. 

The training hyperparameters is as follows: In all three data orderings (impl-first, fact-first, joint), we ensure that every document (i.e.~a fact or an implication) is seen exactly 8 times. In the joint order, all documents are shuffled randomly and trained on for 8 epochs, reshuffling after every epoch. In the fact-first order, we train on facts for 8 epochs, before resetting the optimizer and training on implications for 8 epochs. Impl-first is similar. Note that in Fig.~\ref{fig:data ordering}, the x-axis is normalized to match the number of training steps across the three data orders.

In all settings, we use training hyperparameters described in Sec.~\ref{apx: training details}.

\section{Localizing extractive structures in weight grafting} \label{apx: localizing weight graft}
\begin{figure}[!ht]
    \centering
    \includegraphics[width=0.5\linewidth]{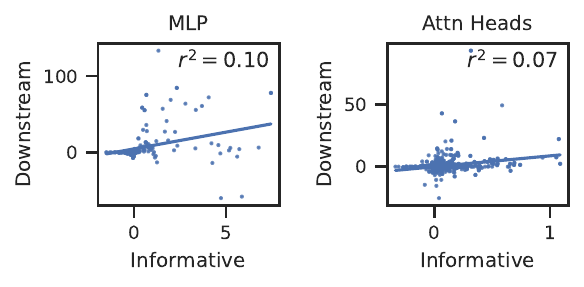}
    \caption{Correlation between downstream extractive scores and informative scores for MLPs (left) and attention heads (right).}
    \label{fig:grafting correlations}
\end{figure}

Earlier in Sec.~\ref{sec: weight grafting} we showed that a certain weight change contained newly learned extractive structures. In this section, we aim to identify the components that are modified by this weight change, and show that they are exactly the downstream extractive structures.

To identify the components modified by the weight change, we use the informative score. Consider the model finetuned on counterfactual train facts $\weights_{\facts F_{\text{train}}'}$, and a reward defined based on how much the model knows the counterfactual train implications $\impl \facts F_{\text{train}}'$. Initially, the reward is low because the model only knows the counterfactual train facts $\facts F'$ but not the train implications $\impl \facts F_{\text{train}}'$. Suppose that after grafting, the model $\weights_{\text{graft}}$ can now predict counterfactual facts $\facts F_{\text{train}}'$ so that the reward is now high. Then, the informative scores with respect to this change in weights $\weights_{\text{graft}} - \weights_{\facts F_{\text{train}}'}$ would identify the important components modified by the weight change, because the informative scores identify components that, under the weight change, contribute to increasing the reward.

To show that these identified components are the downstream extractive structures, we apply the downstream scores to a different weight change. This time, consider the grafted model $\weights_{\text {graft}}$. If we finetune this weights on test facts $\facts F_{\text{test}}$, then the resulting model should perform well on the test implications $\facts F_{\text{test}}$. The downstream score of this finetuning process should capture the downstream extractive structures that predicts corresponding animals from the cities recalled latently.

We compute the informative and downstream scores for the two weight changes described, and correlate the two approaches for localizing them (Fig.~\ref{fig:grafting correlations}). We find that for both MLP components and attention heads, there is a statistically significant positive correlation between the two metrics. However, the Pearson correlation is small, suggesting that the two metrics do not align perfectly.

\section{Learning rate sensitivity} \label{apx: lr sweep}

\begin{figure}[!h]
    \includegraphics[width=\textwidth]{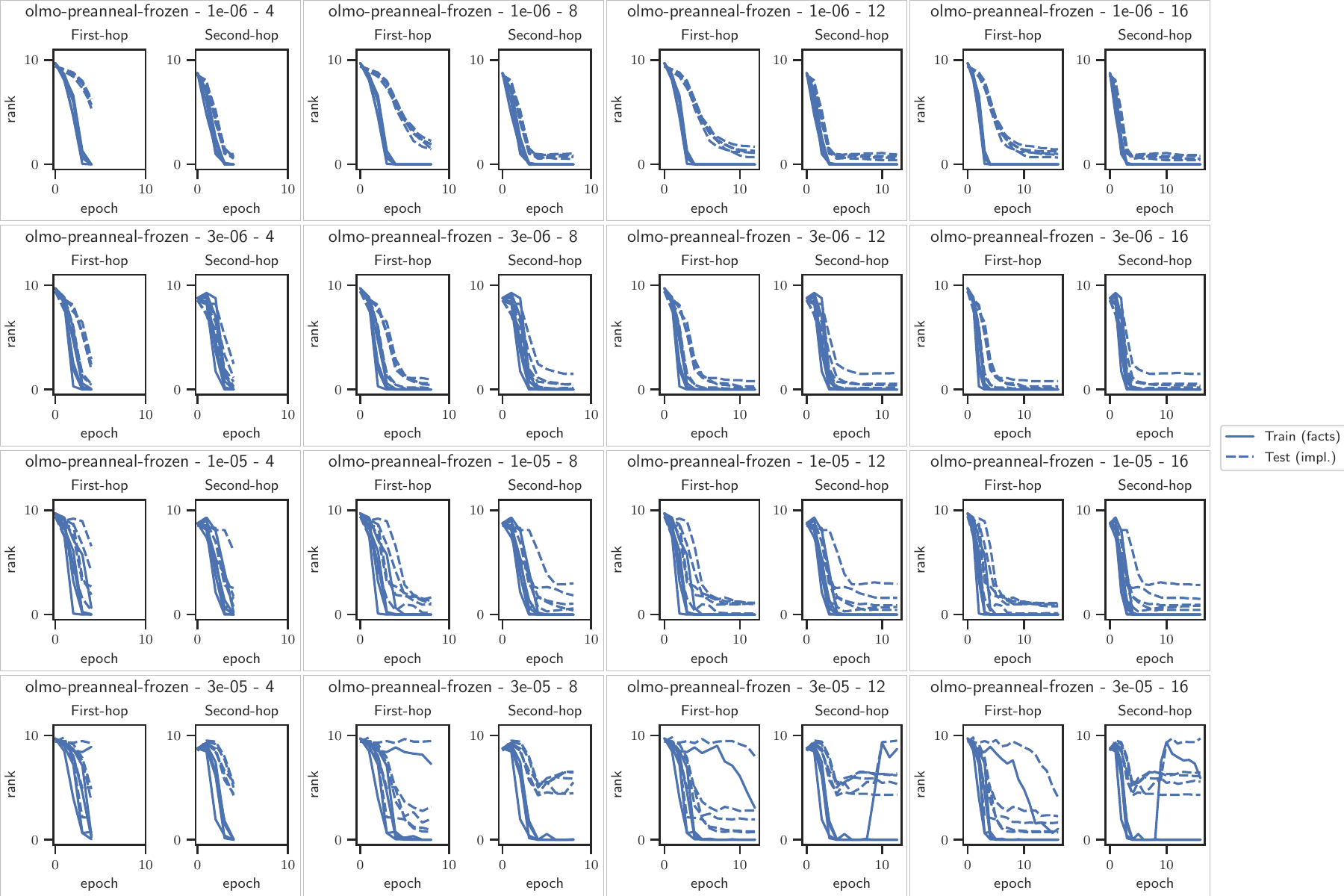}
    \caption{Two-hop OCR performance across different learning rates and epochs for the OLMo-7B pre-anneal checkpoint. 5 random seeds were used. The title has the format ``model - lr - epochs''}
    \label{fig: sweep olmo two hop}
\end{figure}

\begin{figure}[!h]
    \includegraphics[width=\textwidth]{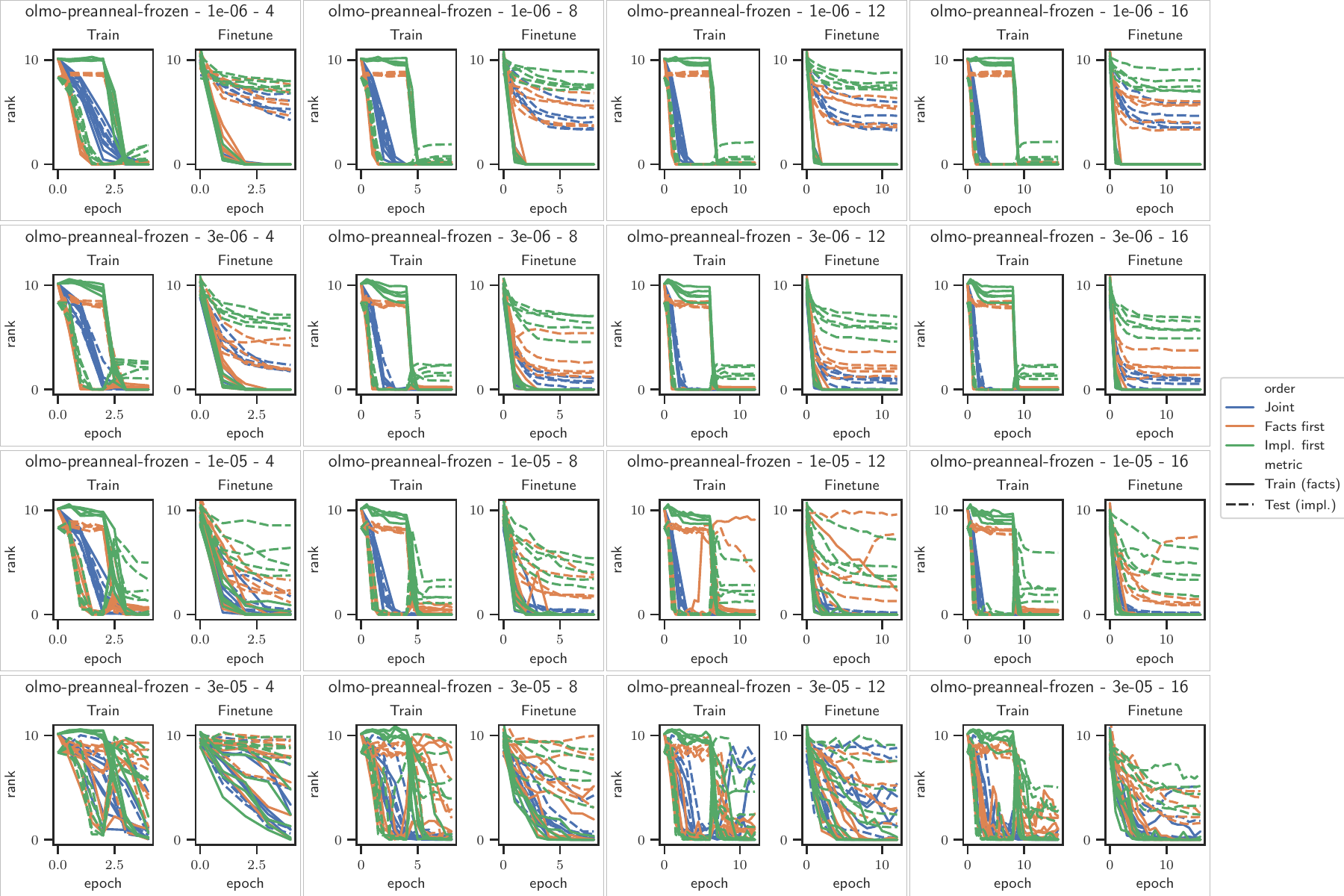}
    \caption{Data ordering effect across different learning rates and epochs for the OLMo-7B pre-anneal checkpoint. 5 random seeds were used. The title has the format ``model - lr - epochs''}
    \label{fig: sweep olmo data}
\end{figure}
\begin{figure}[!h]
    \includegraphics[width=\textwidth]{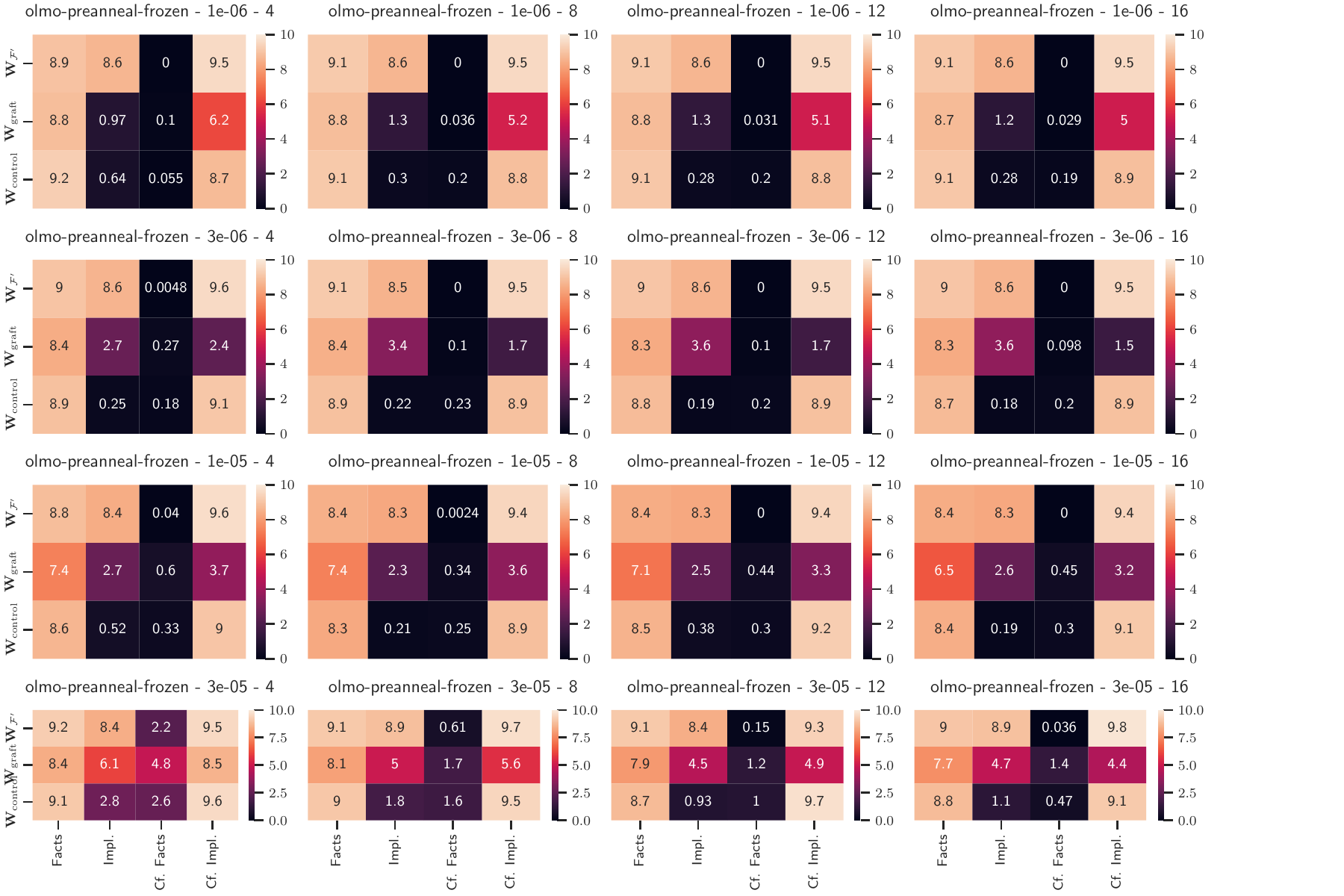}
    \caption{Weight grafting effect across different learning rates and epochs for the OLMo-7B pre-anneal checkpoint. 5 random seeds were used. We show the average mean rank across the 5 random seeds. The title has the format ``model - lr - epochs''}
    \label{fig: sweep olmo weight}
\end{figure}

We observe that the OCR ability exhibits learning rate sensitivity (Fig.~\ref{fig: sweep olmo two hop}). For both learning rates 1e-5 and 3e-6, we observe that the model appears to converge in train rank, but the test rank for 1e-5 is significantly higher than for learning rate 3e-6.

Similarly, data ordering and weight grafting effects are sensitivity to learning rates (Fig.~\ref{fig: sweep olmo data}, \ref{fig: sweep olmo weight}). Further, because whether or not extractive structures are learned in continued pretraining depends on the number of epochs of continued pretraining, the OCR performance also depends strongly on the number of epochs. Nonetheless, in most settings we do observe the expected qualitative pattern: that `Impl-first' data ordering is worse than `Fact-first' or `Joint' at OCR, and that $\weights_{\text{graft}}$ transfers counterfactual implications but not the control graft $\weights_{\text{control}}$.

\section{Other models} \label{apx: other models}
In this section, we show three plots for three models. The three plots are on two-hop OCR performance, data ordering effect, and weight grafting effect. The three models are Llama-3-8b, Gemma-2-9B, and Qwen-2-7B. As in Sec.~\ref{apx: lr sweep}, we sweep across learning rates and epochs, and try 5 random seeds.

Overall, the other models exhibit the same qualitative effects, but the effects are weaker and more sensitive to hyperparameters than in OLMo. We believe that it will be interesting to analyze properties of the models make them less good at OCR.

\begin{figure}[!h]
\begin{subfigure}[b]{1\textwidth}
    \includegraphics[width=\textwidth]{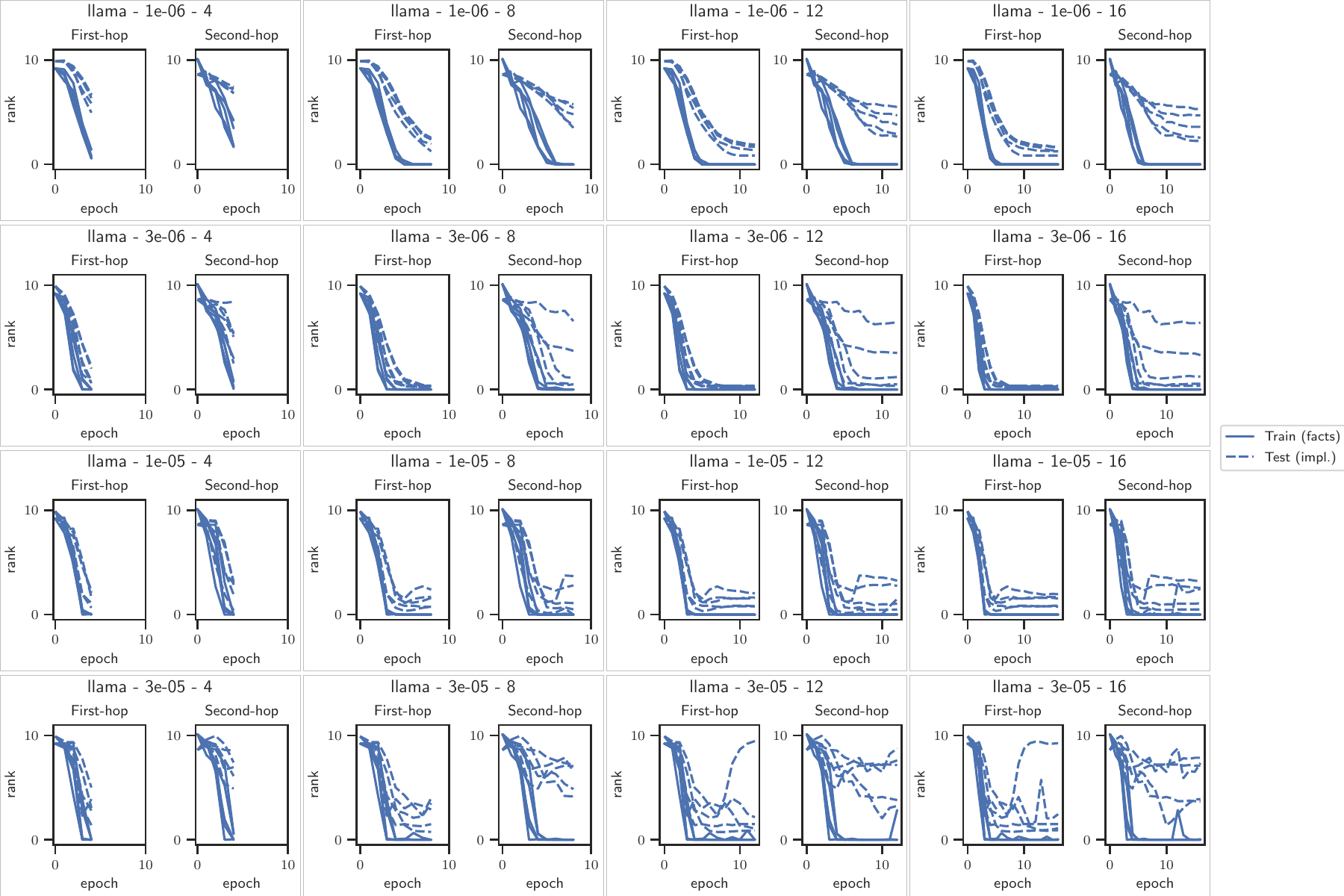}
\end{subfigure}
    \caption{Two-hop OCR performance across different learning rates and epochs for Llama-3-8b. 5 random seeds were used. The title has the format ``model - lr - epochs''}
    \label{fig: llama two hop}
\end{figure}

\begin{figure}[!h]
\begin{subfigure}[b]{1\textwidth}
    \includegraphics[width=\textwidth]{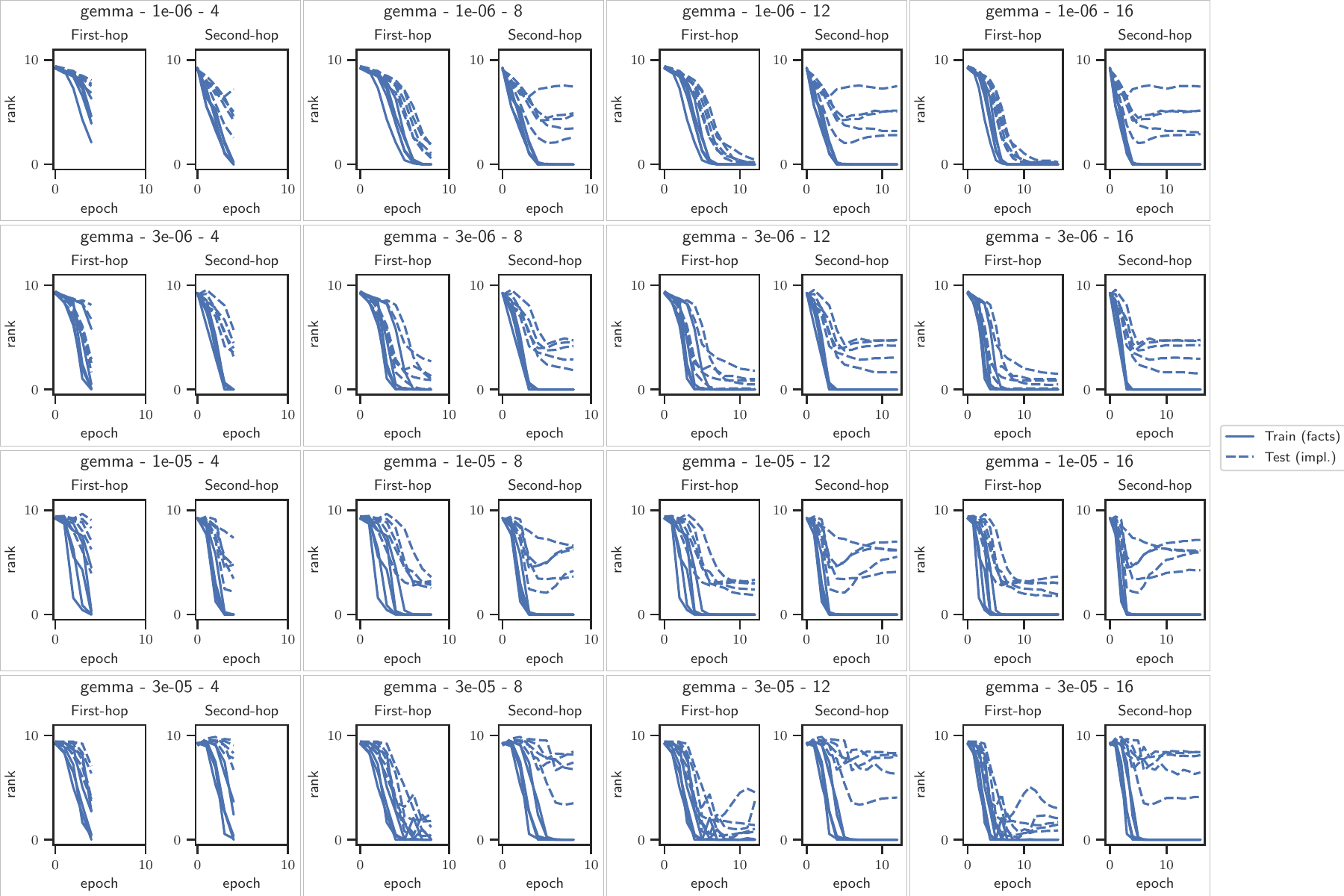}
\end{subfigure}
    \caption{Two-hop OCR performance across different learning rates and epochs for Gemma-2-9B. 5 random seeds were used. The title has the format ``model - lr - epochs''}
    \label{fig: gemma two hop}
\end{figure}

\begin{figure}[!h]
\begin{subfigure}[b]{1\textwidth}
    \includegraphics[width=\textwidth]{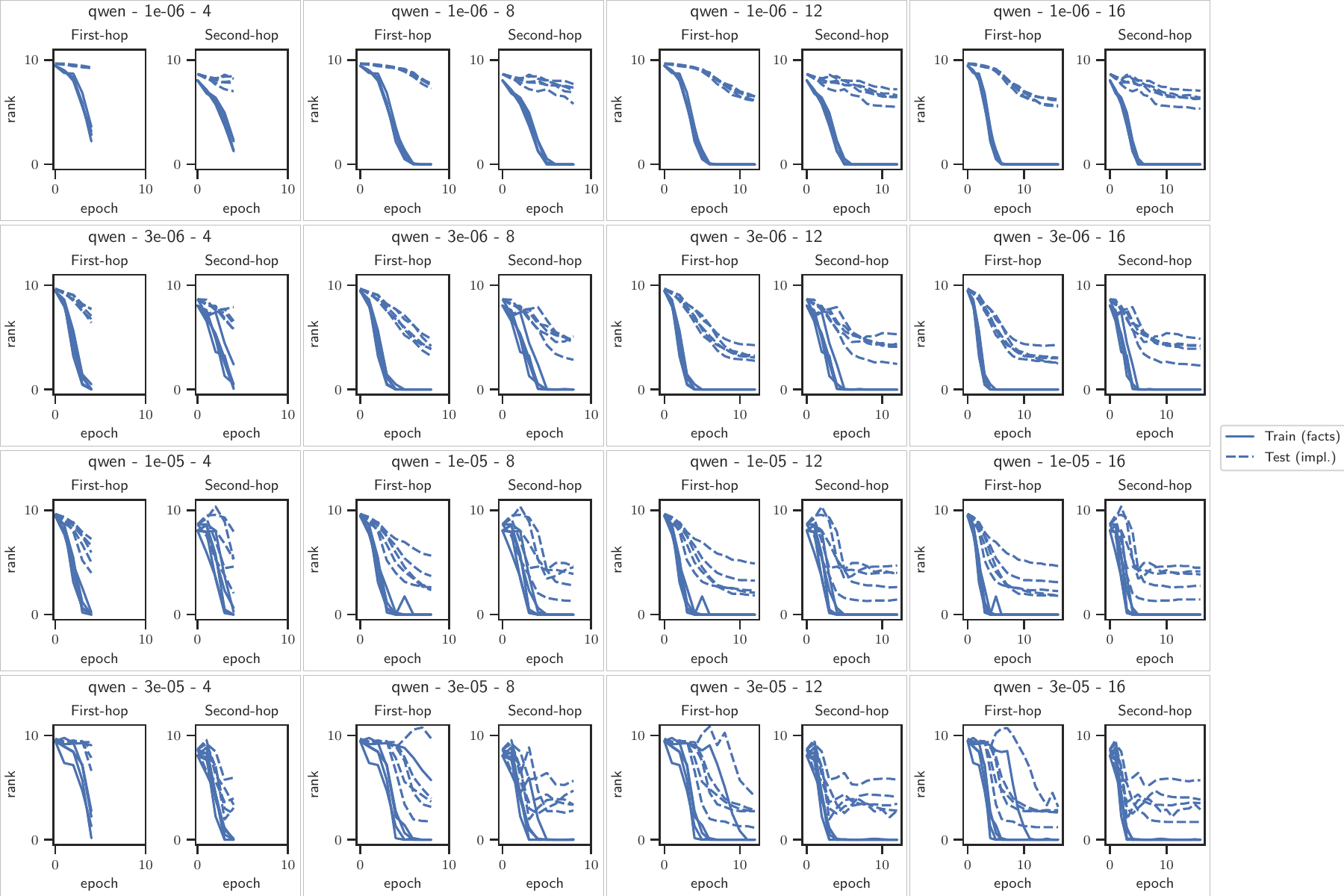}
\end{subfigure}
    \caption{Two-hop OCR performance across different learning rates and epochs for Qwen-2-7B. 5 random seeds were used. The title has the format ``model - lr - epochs''}
    \label{fig: qwen two hop}
\end{figure}

\myparagraph{Two-hop OCR} (Figs.~\ref{fig: llama two hop}, \ref{fig: gemma two hop}, \ref{fig: qwen two hop}) We find that different models require different learning rates to exhibit OCR. All models can perform OCR on the \textsc{First-hop} dataset on some learning rate values. Performance is worse on \textsc{Second-hop} dataset, but the mean rank does decrease significantly across all seeds.

\begin{figure}[!h]
\begin{subfigure}[b]{1\textwidth}
    \includegraphics[width=\textwidth]{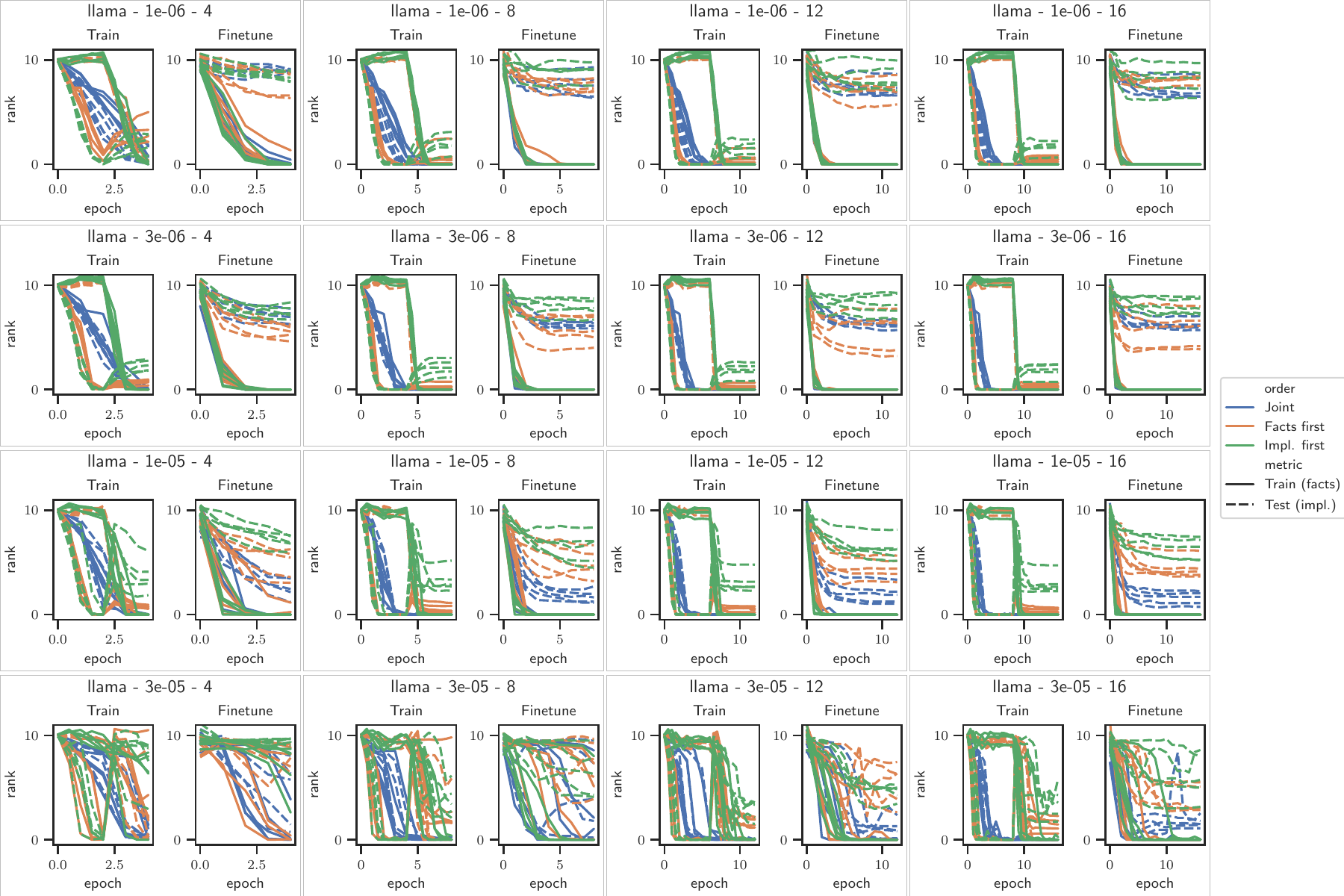}
\end{subfigure}
    \caption{Data ordering effect across different learning rates and epochs for Llama-3-8b. 5 random seeds were used. The title has the format ``model - lr - epochs''}
    \label{fig: llama data}
\end{figure}

\begin{figure}[!h]
\begin{subfigure}[b]{1\textwidth}
    \includegraphics[width=\textwidth]{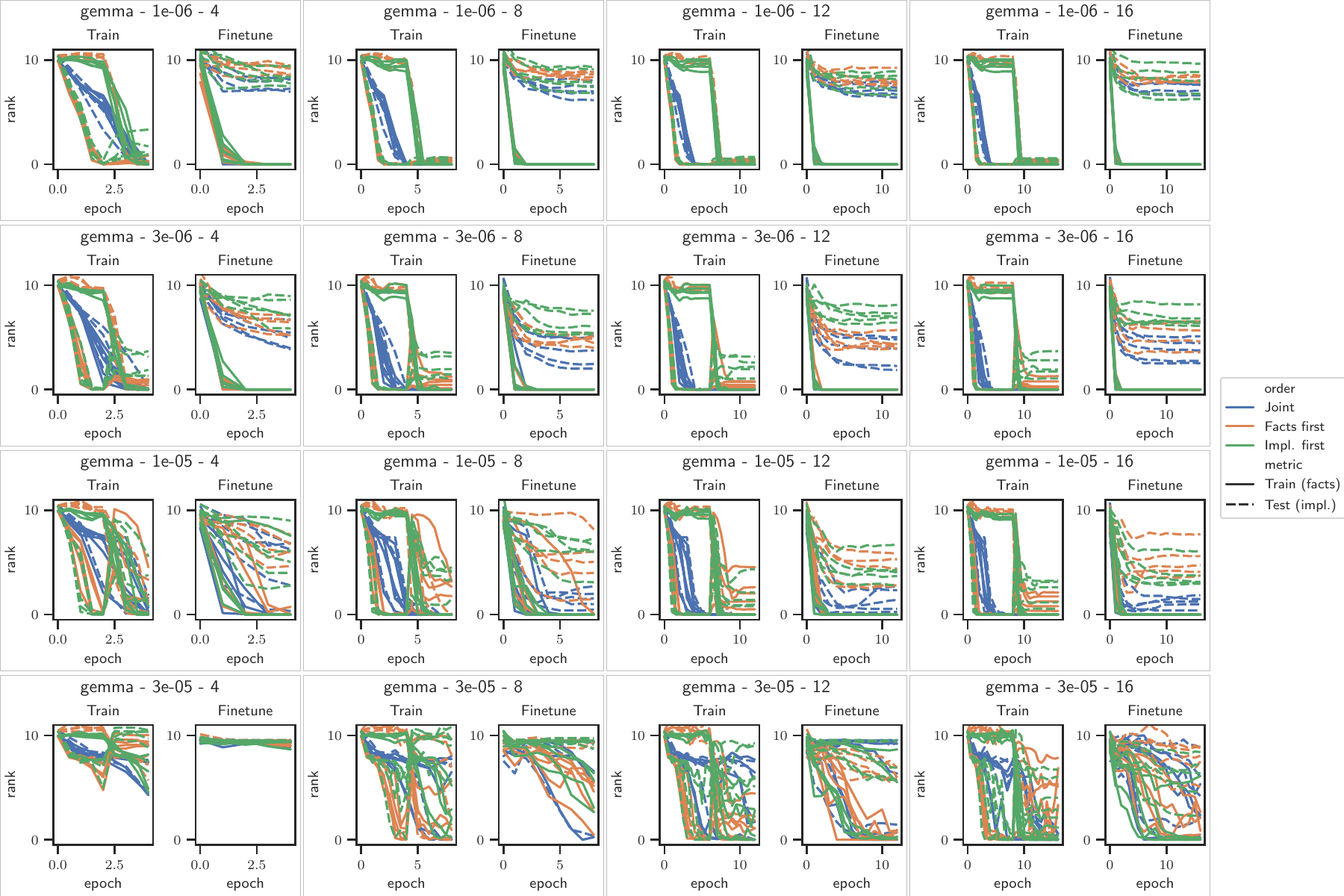}
\end{subfigure}
    \caption{Data ordering effect across different learning rates and epochs for Gemma-2-9B. 5 random seeds were used. The title has the format ``model - lr - epochs''}
    \label{fig: gemma data}
\end{figure}

\begin{figure}[!h]
\begin{subfigure}[b]{1\textwidth}
    \includegraphics[width=\textwidth]{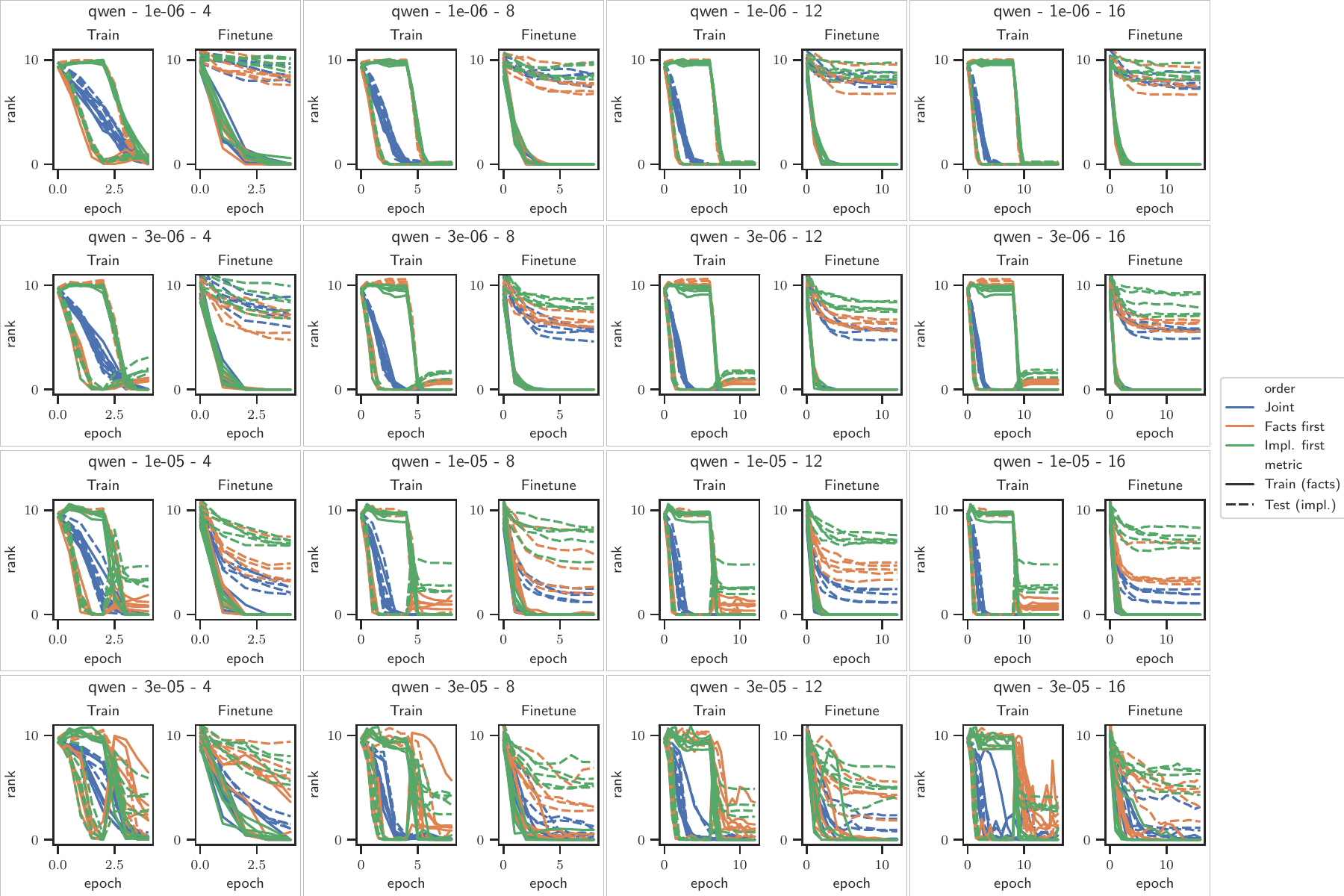}
\end{subfigure}
    \caption{Data ordering effect across different learning rates and epochs for Qwen-2-7B. 5 random seeds were used. The title has the format ``model - lr - epochs''}
    \label{fig: qwen data}
\end{figure}

\myparagraph{Data ordering} (Figs.~\ref{fig: llama data}, \ref{fig: gemma data}, \ref{fig: qwen data}) We observe that `Fact-first' exhibits greater or similar OCR compared with `Impl-first', suggesting that the data ordering effect is present. For some settings, both `Fact-first' and `Impl-first' data orderings have similar, but non-trivial, OCR performance, which again points at the existence of an unknown alternate mechanism for learning extractive structures.

\begin{figure}[!h]
\begin{subfigure}[b]{1\textwidth}
    \includegraphics[width=\textwidth]{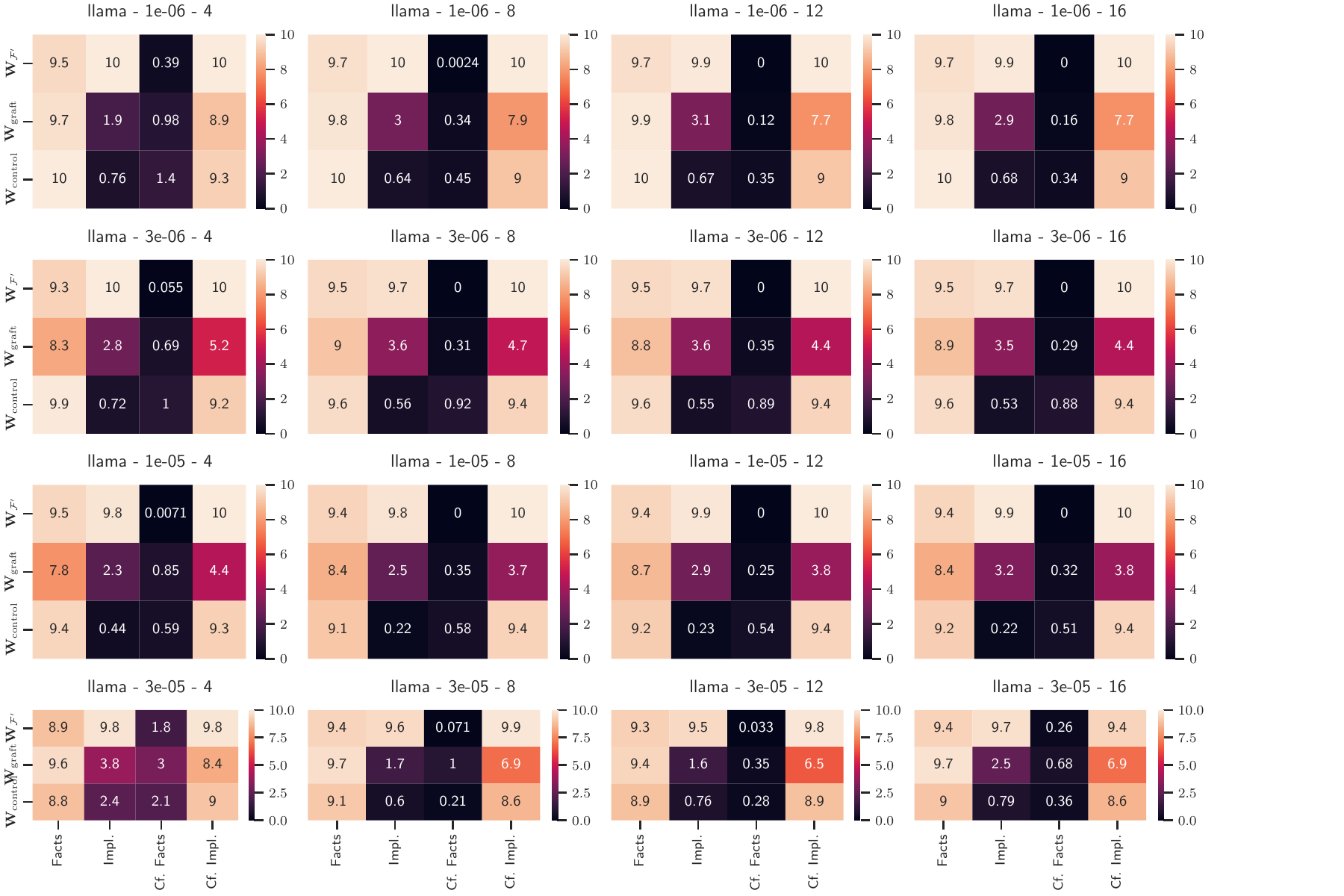}
\end{subfigure}
    \caption{Weight grafting effect across different learning rates and epochs for Llama-3-8b. 5 random seeds were used. The title has the format ``model - lr - epochs''}
    \label{fig: llama weight}
\end{figure}

\begin{figure}[!h]
\begin{subfigure}[b]{1\textwidth}
    \includegraphics[width=\textwidth]{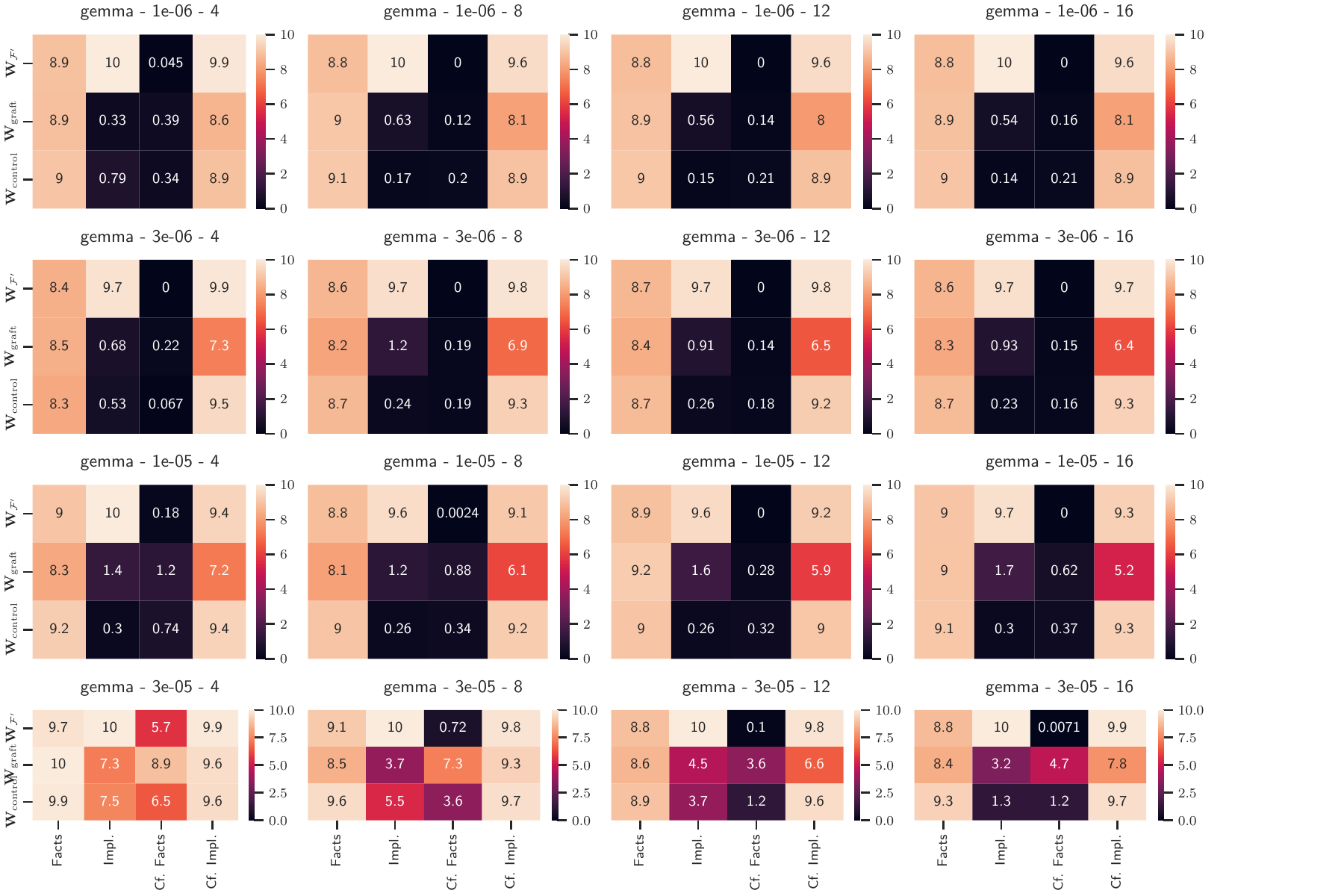}
\end{subfigure}
    \caption{Weight grafting effect across different learning rates and epochs for Gemma-2-9B. 5 random seeds were used. The title has the format ``model - lr - epochs''}
    \label{fig: gemma weight}
\end{figure}

\begin{figure}[!h]
\begin{subfigure}[b]{1\textwidth}
    \includegraphics[width=\textwidth]{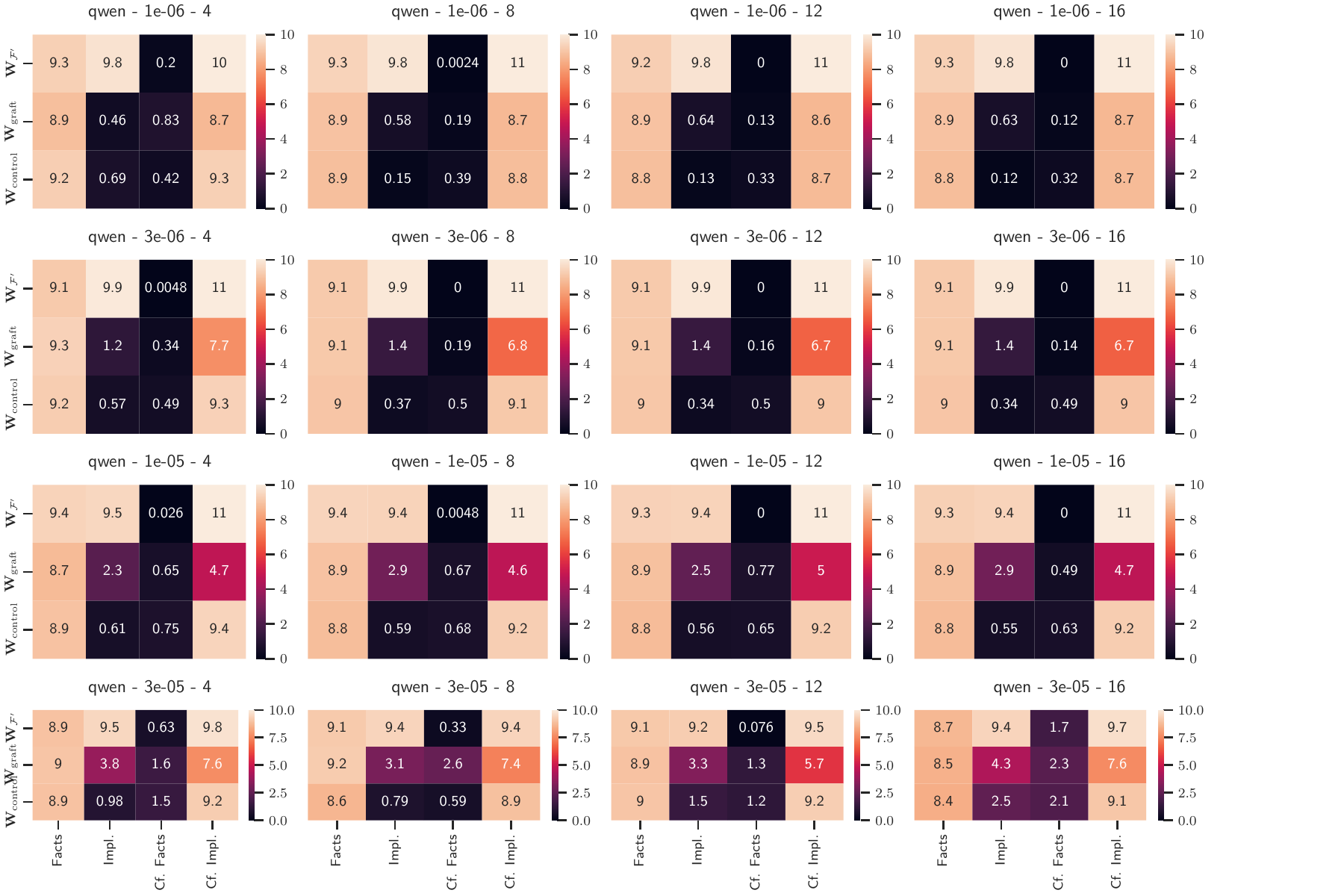}
\end{subfigure}
    \caption{Weight grafting effect across different learning rates and epochs for Qwen-2-7B. 5 random seeds were used. The title has the format ``model - lr - epochs''}
    \label{fig: qwen weight}
\end{figure}

\myparagraph{Weight grafting} (Figs.~\ref{fig: llama weight}, \ref{fig: gemma weight}, \ref{fig: qwen weight}) All models, under the weight graft $\weights_{\text{graft}}$, for an appropriate choice of hyperparameters, can predict counterfactual implications. However, in these models the weight graft also transfers the original implications, reinforcing the hypothesis that these models perform OCR less effectively, and instead memorize more.




\end{document}